\documentclass{article}

\usepackage{ijcai22}

\usepackage[utf8]{inputenc} 
\usepackage[T1]{fontenc}    
\usepackage{hyperref}       
\usepackage{url}            
\usepackage{booktabs}       
\usepackage{amsfonts}       
\usepackage{nicefrac}       
\usepackage{microtype}      
\usepackage{xcolor}         
\usepackage{graphicx}

\usepackage[ruled]{algorithm2e}
\usepackage{multirow}
\usepackage{makecell}
\usepackage{graphicx}
\usepackage{float}
\usepackage{subfig}
\usepackage{amssymb}

\title{FedEntropy: Efficient Device Grouping for Federated Learning Using  Maximum Entropy Judgment}

\author{
 Zhiwei Ling,
 Zhihao Yue,
 Jun Xia,
 Ming Hu,
 Ting Wang,
 Mingsong Chen$^*$\\
 $^1$Shanghai Key Lab of Trustworthy Computing, East China Normal University\\
$^*$Corresponding Author, \{mschen\}@sei.ecnu.edu.cn
}

\begin{document}

\maketitle

\begin{abstract}
Along with the popularity of Artificial Intelligence (AI) and Internet-of-Things (IoT), Federated Learning (FL) has attracted steadily increasing
attentions as a promising  distributed machine learning paradigm, which enables
the training of  a central model on for numerous
decentralized devices  without exposing their
privacy. However,  due to the biased  data distributions on involved
devices,   FL inherently suffers from  low classification
accuracy in non-IID scenarios. 
Although  various  device grouping method have been 
proposed to address this problem, most of them neglect both 
i) distinct data distribution  characteristics of heterogeneous devices,  and
ii) contributions and   hazards of local models, which are extremely 
important  in determining the quality of global model aggregation. 
In this paper, we present an effective FL method named FedEntropy
with a novel dynamic
device grouping scheme, 
which makes full use of the above two factors based on 
our  proposed maximum entropy judgement heuristic. 
Unlike existing FL methods that directly
aggregate local models returned from all the selected devices, in one FL round
FedEntropy firstly 
makes a  judgement based on the pre-collected
soft labels of selected
devices and then only aggregates the local models that can 
maximize the overall entropy of these soft labels. 
Without collecting local models that are harmful for 
aggregation, FedEntropy can effectively improve  
global model  accuracy while reducing the overall communication overhead. 
Comprehensive experimental results on well-known benchmarks show that,
FedEntropy not only   outperforms
 state-of-the-art FL methods in terms of  model accuracy and communication overhead, but also can be integrated into them to enhance their classification
 performance.   
\end{abstract}

\section{Introduction}
Due to the prosperity of Artificial Intelligence
(AI) and Internet of Things (IoT), Federated Learning (FL) \cite{mcmahan2017communication,GhoshCYR20,MitraJPH21}
has been increasingly used in various
security-critical AI IoT (AIoT) applications (e.g., autonomous driving, 
commercial surveillance, and healthcare informatics). Unlike traditional 
centralized machine learning  methods, 
FL provides a decentralized   framework that can train a 
global Deep Neural Network (DNN)
model without compromising
user  privacy (i.e., data) distributed on devices. 
Typically, FL adopts a cloud-device architecture, where the cloud manages a 
global model for the knowledge fusion of local models on
devices. During the FL training, devices only need to send their local models to 
the cloud for aggregation. Without sending local device data directly to 
the cloud, the data privacy of devices can be guaranteed \cite{NEURIPS2021_285baacb}.

Although FL is good at sharing
knowledge  among 
all the involved 
devices, due to  
biased  device  data distributions in non-IID (Not Independent and Identically Distributed) scenarios, 
it greatly suffers from 
 low classification 
 accuracy caused by 
  statistical heterogeneity of  datasets  and  high communication overhead due to 
 periodic model aggregation \cite{NEURIPS2021_2f2b2656,DBLP:conf/nips/KarimireddyJKMR21,DBLP:conf/aistats/0001MR20}.
To mitigate these problems, 
various  variants have been proposed 
to improve FL performance from different  perspectives, such 
as  constrained local updates \cite{li2020federated,karimireddy2019scaffold,li2021model}, device grouping \cite{DBLP:conf/bigdataconf/ChenCZK20,DBLP:conf/icml/FraboniVKL21}, and knowledge distillation \cite{zhu2021data,DBLP:journals/corr/abs-1910-03581,DBLP:conf/nips/LinKSJ20}. 
Although they can effectively 
increase the classification accuracy or reduce the communication overhead of 
vanilla FL in non-IID scenarios, most of them treat all the devices equally  without taking their specific characteristics and effects into account during the 
aggregation. 
Simply aggregating local models to the cloud in an 
equal manner will inevitably result in
a notable loss of classification accuracy and a slowdown of convergence \cite{DBLP:journals/corr/abs-2105-07066,DBLP:conf/mm/Zhuang0ZGYZZY20}.
In this case, some useless local models are forced to 
participate the aggregation in one FL training round, which may affect the 
aggregation quality as well as result in more communication overhead. Worse still, 
a  participant device may upload a harmful  local model
with extremely
biased data distribution
to the cloud, which may significantly
degrade the aggregation quality resulting in 
low classification accuracy. 
Therefore, {\it how to accurately model the 
distribution characteristics  of device data
and utilize them to facilitate 
the global model aggregation to achieve better 
overall performance is becoming a
challenge in FL design}.


Since entropy can accurately quantify the 
information carried by a specific probability distribution, it 
has been widely used for various classification purposes
 \cite{DBLP:conf/nips/DubeyGRN18}.
Based on the merits of entropy,  in this paper we
propose an effective FL method named FedEntropy with 
a novel
dynamic device grouping scheme, which enables the 
 evaluation and selection of device models before 
they are uploaded to the cloud. 
Unlike traditional FL, in one training round of
FedEntropy, the cloud firstly collects
the soft labels from selected 
devices and judges  their
contributions to the overall entropy 
of soft labels. Then, only a subset of 
selected devices that  maximize the overall 
entropy  need to upload their local models to the cloud
for aggregation. 
In this way, FedEntropy effectively and safely 
filters both useless and harmful devices for 
aggregation, thus leading to higher classification 
performance with less communication overhead. 
This paper makes the following three contributions:
\begin{itemize} 
\item  
Based on the concept of maximum entropy, 
we propose an effective 
dynamic device grouping method, which
takes the specific data distribution characteristics of devices and 
their contributions to 
the aggregation into account. 

\item
We introduce a novel  FL framework that enables fine-grained
device selection and corresponding model aggregation
based on our proposed maximum entropy 
judgment method.

\item
We evaluate the performance of our FedEntropy method on 
various classic datasets, and demonstrate both the 
superiority and compatibility of FedEntropy  over state-of-the-art FL methods in terms of classification accuracy and communication overhead. 

\end{itemize}

The rest of this paper is organized as follows. 
Section \ref{related_work} briefly reviews the related work. Section \ref{approach_design} details the design of our proposed FedEntropy method. Section \ref{experiments} presents the experimental results. Finally, Section \ref{conclusion} concludes the paper.

\section{Related Work}
\label{related_work}

To improve the classification accuracy in non-IID scenarios, 
existing FL methods can be classified into three major 
categories based on  local training correction, 
device grouping, and knowledge distillation, respectively. 
The 
{\bf local training correction-based methods} 
  require modifying  loss functions during local model training. 
For example, by adding a proximal term to regularize the local loss function, Li et al. \cite{li2020federated} introduced FedProx that
  can make  trained local models get closer to the global model. Based on model-contrastive learning, Li et al. proposed 
Moon \cite{li2021model}, which  can 
correct the deviation of  local models
from their global model.
In \cite{karimireddy2019scaffold}, 
Karimireddy et al. presented a stochastic controlled averaging scheme
called SCAFFOLD, which can address the ``client drift'' issue caused by data heterogeneity.
However, the introduction of 
 global control variables in SCAFFOLD incurs 
 high communication overhead, which  is twice than that of FedAvg \cite{mcmahan2017communication}.
The  {\bf device grouping-based approaches} take the data similarity among all devices into account. 
For example, Duan et al. \cite{DBLP:journals/tpds/DuanLCLTL21} 
proposed Astraea, which groups local models
based on the KL divergence of their data distributions and builds mediators to reschedule the training of local models.
Fraboni et al. \cite{DBLP:conf/icml/FraboniVKL21} developed
a method for clustering local models
based on sample size or model similarity, which can improve the representativity of local models  and decrease the variance of random aggregation in FL.
The {\bf knowledge distillation-based approaches} use soft labels generated by a ``teacher model'' to guide the training of ``student models''. 
For example,  Zhu et al. 
\cite{zhu2021data} presented
a data-free knowledge distillation method called FedGen that uses a lightweight generator to correct local training.
Under the help of some
unlabeled dataset, Lin et al. \cite{DBLP:conf/nips/LinKSJ20} 
proposed FedDF, which accelerates the training convergence by 
adopting  outputs of local models to train 
the global model. 
 Although the above FL
 methods are promising, they do not take the data distribution characteristics of devices into account, which can be used 
 to further improve the 
 classification performance in non-IID scenarios.


As an essential principle of Bayesian 
statistics,  Maximum Entropy Principle (MEP) \cite{jaynes1957information}
 has been widely applied in many areas of machine learning, including supervised learning
 \cite{DBLP:journals/jmlr/ChenGGRC09,DBLP:conf/nips/ZhuXZ08}
 and reinforcement learning.  
For testable information,  MEP assumes
that the probability distribution 
best representing the current state of knowledge
is the one with the largest entropy, where
the distribution should be as uniform as possible. 
In  supervised learning, most of 
existing MEP-based methods \cite{DBLP:journals/jmlr/Shawe-TaylorH09,DBLP:conf/iclr/PereyraTCKH17}
focus on the entropy-based optimization
of  classifiers  to 
improve classification accuracy.
Our approach is inspired by the work in 
 \cite{DBLP:conf/nips/DubeyGRN18},
 which applies MEP on soft labels generated by classifiers. However, none of them 
 take MEP into account to optimize the 
 classification performance in FL.
To the best of our knowledge, our work is the first attempt  that  applies 
soft label-based 
MEP on device grouping. 
Since our approach  makes full use of device data distribution characteristics in non-IID scenarios, it can not only enhance the overall FL classification performance, but also reduce the communication overhead between the cloud and devices.

\begin{figure*}[th]
  \centering
  \includegraphics[width=3.5in]{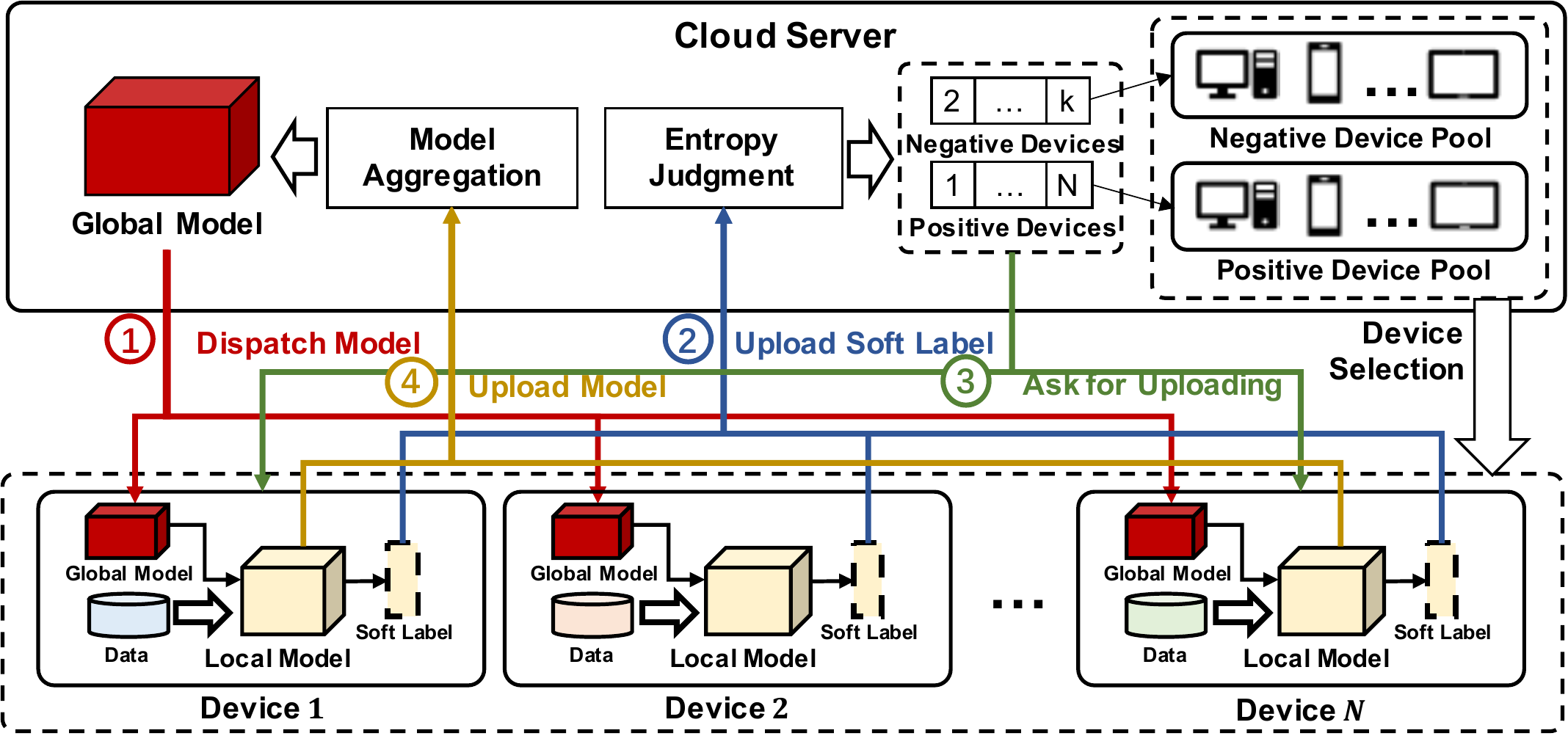}
  \caption{Architecture and workflow of our FedEntropy framework.}
  \vspace{-0.2in}
  \label{f1}
\end{figure*}

\section{Our FedEntropy Approach}
\label{approach_design}

\subsection{Preliminaries and Problem Formulation}

Typically, FL adopts a cloud-device architecture, where the cloud manages a 
central global model to aggregate the knowledge uploaded from devices. 
Assume that in an FL system there exist   $N$  devices and $l$ samples in total,
where the dataset 
on the $k^{th}$ device is $D_k$, and the number of samples on 
the $k^{th}$ device is $l_k$  (i.e., $|D_k|=l_k$) such that $l = \sum_{k=1}^N l_k$.
Let $(x_k^i,y_k^i)$ be the $i^{th}$ sample on the $k^{th}$ device. 
Let  $w$ and $w_k$ be the models on the server and the client, respectively, and 
$\Phi$ be the user-defined loss function for devices.
Similar to traditional FL, the  objective of our approach 
is formulated as follows:
\begin{scriptsize}
\begin{equation}
    \mathop{\min}\limits_{w} \Big\{F(w) \triangleq \sum_{k=1}^N \frac{l_k}{l} f_k(w)\Big\}, \\
    \quad f_k(w) \triangleq \frac{1}{l_k} \sum_{i=1}^{l_k} \Phi(w;(x_k^i, y_k^i)).
\end{equation}
\end{scriptsize}


\subsection{Overview of FedEntropy}

Different from traditional FL methods that  treat  all the involved devices
in an equal manner, our approach claims
that all the devices
in non-IID scenarios are different due to their unique
data distribution characteristics.  Based on our observation, it is not suitable  to randomly group  local models on devices for the purpose of model aggregation, since 
 contradictive  local models within a same group  may interfere with each other, resulting in low classification accuracy. To avoid this case, in FedEntropy we adopt a two-stage
 device selection strategy to 
 enable fine-granularity 
 device grouping. In the first stage, FedEntropy collects the soft labels from 
 a coarse set of selected devices rather than their local models, indicating the 
 data distributions on devices. 
 By using our 
 proposed maximum entropy judgement heuristic,  FedEntropy can 
 quickly figure out which 
 local models are not friendly for the aggregation based on the collected 
 soft labels and filter their corresponding devices 
 away from the group. Based on the fine-tuned 
 new device group, the second stage acts in the same way as the traditional FL methods.

 Figure \ref{f1} details both  the architecture of   and the  
 workflow for FedEntropy, where one  training round of FedEntropy
 involves four steps.  
Unlike transitional FL, on the cloud  FedEntropy maintains two  pools of device indices, 
i.e., {\it positive device pool} and  {\it negative
device pool}, where initially the  positive device pool contains all the devices. 
At the beginning of a training round, the cloud server randomly selects 
 partial (i.e., with the number of 
 $N \times C$ where $0 \leq C \leq 1$)   devices 
 from either positive pool or negative pool 
 according to the $\epsilon$-greedy policy. 
In step 1 the cloud  broadcasts the global model to 
 all the selected devices. Once the selected devices finish
 the local training, in step 2 they only forward their soft labels
 to the server. Note that,
 compared with the size of a local model, 
 the size of a soft label is negligible. 
 All the collected soft labels  will then be 
 processed by the maximum 
 entropy judgment module of the cloud, 
 where  labels that lead to  maximum  aggregated information entropy
 will be used for  fine-grained device grouping. For all the selected 
 devices in this training round, we 
mark  the devices whose soft labels
 contribute to the maximum   entropy
as {\it positive devices}, while the other remaining
devices are marked as 
  {\it negative devices}. 
 In step 3, only the positive devices need to upload their
 local models for the aggregation, which will be executed in step 4.  
At the end of each round, the  positive devices and negative devices
will be put into the {\it positive device pool} and  {\it negative
device pool},  respectively.


\subsection{Maximum Entropy Judgment}
\label{3.2}

Our approach  adopts  entropy to judge
the similarity 
between the data distributions of devices. 
 To accurately calculate such information entropy, 
it is required to obtain the probability distributions of  datasets from devices. 
However, if all the devices send their datasets to the cloud, 
the FL assumption of data privacy will be substantially violated \cite{DBLP:conf/infocom/WangSZSWQ19}. 
Instead of sending datasets directly to the cloud, 
our approach adopts the averaged soft label information of samples to implicitly 
represent  the dataset distributions of devices.  In this way, the data privacy of devices can be guaranteed.


Based on the outputs of the softmax layer of a local model, we 
collect all the soft labels
of samples including the ones of  both correct  and incorrect predictions, 
where a soft label indicates the data distribution within a sample. 
We then aggregate  these soft labels together
to form an indicator of the data distribution  of 
all  samples on the same device.  
For a device
with an index of $k$, assuming that it has $l_k$ samples, we calculate the data
distribution
of corresponding device by averaging all its
soft labels  using the formula as follows: 
\begin{equation}
    p_k = \frac{1}{l_k} \sum_{i=1}^{l_k} \phi_k(x_k^i),
    \nonumber
\end{equation}
where $\phi_k(x)$ denotes the  softmax layer
output of the $k^{th}$ local model when the input is $x$. 
Assume that in one FL training round, a set of   devices are selected whose sample distributions form  a set
$\{ p_1^{\prime}, p_2^{\prime}, \ldots, p_n^{\prime} \} $. 
Assuming that there are $c$ data categories   in total for the whole FL training, 
the information  entropy for the probability distribution $p^{\prime}$ is defined as follows:
 \begin{equation}
     Entropy(p^{\prime}) = - \sum_{i=1}^c (p^{\prime}[i] * log(p^{\prime}[i]).
     \nonumber
 \end{equation}

According to the intrinsic property of information entropy, 
only when the  data distribution of selected 
devices is uniform, we can achieve the highest information entropy.
Based on the observation in \cite{DBLP:journals/tpds/DuanLCLTL21}, when the overall
data distributions on the selected devices is uniform (i.e., 
the datasets are complementary to each other), we can achieve the best 
aggregation quality. In other words,  the  higher  the overall   entropy of soft labels sent by 
selected devices is , the better the aggregation quality
we can achieve.
However, it is hard to achieve such a uniform distribution in FL, especially for 
non-IID scenarios. In order to make 
an entropy of some device group as high as possible,
we design a maximum entropy judgement, which can effectively filter the devices that has 
negative impacts on the maximum entropy calculation.

\begin{algorithm}
\caption{Implementation of Maximum Entropy Judgement}
\label{alg:checkentropy}
\LinesNumbered
\KwIn{
i) $\mathcal{P}$,\, list of aggregated soft labels for devices; 
ii) $\mathcal{L}$,\, list of device dataset sizes;
iii) $S_t$,\, set of indices of selected devices;
}
$\mathcal{A} \leftarrow S_t, \mathcal{R} \leftarrow \{\}$\\
\While{$|\mathcal{P}| > 0$}
{   
    $Entropy \leftarrow getEntropy(\mathcal{P}, \mathcal{L})$\\
    $index \leftarrow 0$\\
    \For{each $k \in S_t$}
    {  $\overline{\mathcal{P}}\leftarrow \mathcal{P} $, $\overline{\mathcal{L}}\leftarrow \mathcal{L} $     \\
     $ \overline{\mathcal{P}}$.remove($k$), $ \overline{\mathcal{L}}$.remove($k$)\\
    $\overline{Entropy} = getEntropy(\overline{\mathcal{P}}, \overline{\mathcal{L}})$\\
    \If{$\overline{Entropy} > Entropy$}
    {$Entropy \leftarrow \overline{Entropy}$, $index \leftarrow k$}}
    \eIf{$index == 0$}
    {\textbf{break}}
    {$\mathcal{A} \leftarrow \mathcal{A} \setminus \{index\}, \mathcal{R} \leftarrow \mathcal{R} \cup \{index\}$\\
     $\mathcal{P}$.remove($index$), $\mathcal{L}$.remove($index$)}
}
\Return{$(\mathcal{A}, \mathcal{R})$}
\end{algorithm}

Algorithm \ref{alg:checkentropy} details the implementation of
our maximum entropy judgment module on the cloud. 
Note that the inputs of this algorithms include the 
 aggregated soft labels of  selected devices and their corresponding dataset sizes. 
 In this algorithm, initially
 we use $\mathcal{P}$ to save all the collected   aggregated soft labels.
 Then, our approach 
 iteratively searches for soft labels in $\mathcal{P}$ that can degrade the 
 overall entropy of data distributions of selected devices, until no 
aggregated  soft label in the positive device  set (i.e., $\mathcal{A}$) is harmful to 
the overall entropy. 
Line 1 initializes  $\mathcal{A}$ and $\mathcal{R}$, where $\mathcal{R}$  denotes the set of
devices that are filtered away from   $\mathcal{A}$.
Lines 2-19 iteratively judge whether there exist aggregated soft labels that  are harmful for 
the overall entropy. 
Line 3 recalculates the overall entropy of the aggregated soft labels for positive devices
based on the formula as follows:
\begin{equation}
    getEntropy(\mathcal{P}, \mathcal{L}) = Entropy(\frac{\sum_{i \in \mathcal{A}} (\mathcal{P}[i] \times \mathcal{L}[i])}{\sum_{i \in \mathcal{A}} \mathcal{L}[i]})
    \nonumber
\end{equation}
Lines 4-12 traverse all the aggregated soft labels to detect whether there exists one
that is harmful for the overall entropy. If exist, lines 16-17 will 
remove the corresponding  device from the positive device set and put it to the negative device set. 
Otherwise, line 14 will terminate the search. Finally, line 20 returns 
the judgment results based on the concept of maximum entropy. 




\subsection{Implementation of FedEntropy}

As shown in Figure \ref{f1}, our approach maintains two pools for 
positive devices and negative devices, respectively. Although the devices 
in the positive device pool can contribute more to the overall entropy, it does not
mean the datasets of negative devices are useless. In fact, 
the main reason that a device is in the negative device 
pool is because it fails to  get on well with majority 
of devices involved in FL. 
In order to encourage all the devices to  participate the global model
aggregation, we propose a dynamic device grouping strategy.
At the beginning of each training round, the cloud
uses the $\epsilon$-greedy policy to select
devices from either the positive pool 
(with a probability of $\epsilon$) or the 
negative pool (with a probability of $1-\epsilon$), 
where $\epsilon = 0.8$  by default. 
If the number of devices in one pool is insufficient, 
our approach will randomly select the remaining  devices  from the other 
pool. Based on this cloud setting, our FedEntropy method is implemented as follows. 
  
\begin{algorithm*}
\caption{Inplementation  Details of FedEntropy}
\label{alg:aggregation}
\LinesNumbered
\KwIn{
i) $T$,\, \# of training rounds;\,
ii) $N$,\, \# of total clients;\,
iii) $C$,\, fraction of active clients in each round;\,
iv) $D_k$,\, dataset of $k^{th}$ client;\, 
v) $w_g$,\, parameter of global model;
}
$Initialize(w_g)$ \\
$S_{p} \leftarrow \{1,2,\dots,N\}$, $S_{n} \leftarrow \{\} $ \\
\For{$t=1,\dots,T$} { 
    \eIf{$random(0, 1.0) < \epsilon$}
    {$S_t \leftarrow$ $RandomSet(S_p,N,C,\epsilon)$,  $S_{p} \leftarrow S_{p} \setminus S_t$}
    {$S_t \leftarrow$ $RandomSet(S_n,N,C,1-\epsilon)$,, $S_{n} \leftarrow S_{n} \setminus S_t$}
    $\mathcal{P} \leftarrow [\ ], \mathcal{W} \leftarrow [\ ], \mathcal{L} \leftarrow [\ ] $\\
    \For{$k \in S_t$ \textbf{in\ parallel}}{
        $(p_k, l_k) \leftarrow  ClientUpdate(w_g, D_k)\quad$ 
        \tcp{Local training on the $k^{th}$ device}
        $\mathcal{P}[k] \leftarrow p_k, \mathcal{L}[k] \leftarrow l_k$
    }
    $(\mathcal{A}, \mathcal{R})\leftarrow JudgeEntropy(S_t, \mathcal{P}, \mathcal{L})$\\
    \For{$k \in S_t$}{
        \If{$k \in \mathcal{A}$}
        {
        $w_k \leftarrow$ local model of $k^{th}$ device
        \\
        $\mathcal{W}[k] \leftarrow w_k$}
    }
    $w_g \leftarrow \frac{\sum_{i \in \mathcal{A}}(\mathcal{L}[i]*\mathcal{W}[i])}{\sum_{i \in \mathcal{A}}\mathcal{L}[i]}\quad$\tcp{Aggregation}
    $S_{p} \leftarrow S_{p} \cup \mathcal{A}, S_{n} \leftarrow S_{n} \cup \mathcal{R}$ 
} 
\Return{$w_g$}
\end{algorithm*}

Algorithm \ref{alg:aggregation} details
the interactions between the cloud server 
and devices during the training of  FedEntropy, where lines 3-23 conduct 
$T$ rounds of FedEntropy training. 
Based on the $\epsilon$-greedy policy, 
lines 4-8 select  a set   devices from either  
 the positive pool or the negative pool. 
Line 9 initializes   $\mathcal{P}$, $\mathcal{W}$, and $\mathcal{L}$, 
which are used for storing 
corresponding
soft labels, models, and dataset sizes of selected 
devices, respectively. 
In lines 10-13, 
the cloud broadcasts the global model to all the selected devices in 
$S_t$ to trigger  local training by the devices in parallel.
After collecting  all the aggregated labels from 
 selected devices, line 14 figures out both the 
 positive  devices and negative devices for the aggregation
 using the function \textit{JudgeEntropy} implemented in Algorithm \ref{alg:checkentropy}.
In lines 15-20, the cloud asks
the positive devices in $\mathcal{A}$ to upload their local models
for the aggregation in line 21. 
Meanwhile, both positive and negative device pools are updated 
 in line 22. Finally, the algorithm achieves a trained global model for all the involved devices in some non-IID scenario.

\noindent{\bf Convergence Analysis and Discussions}.
Similar to \cite{DBLP:journals/corr/abs-2203-09249}, we use 
FedAvg as the FL optimizer of FedEntropy. 
As stated in \cite{li2019convergence}, FedAvg converges at a rate of $O(\frac{1}{t})$. 
Since FedEntropy selectively aggregates a portion of devices per round,
we can reasonably regard FedEntropy as a variant of FedAvg 
that aggregates  a subset of selected
devices per round. Therefore, 
both FedEntropy and FedAvg have the same convergence rate, i.e.,  $O(\frac{1}{t})$.
Note that, FedEntropy can also adopt other existing FL methods as its 
optimizer, where
the convergence rate of FedEntropy is 
the same as theirs. 




\section{Experiments}
\label{experiments}

To evaluate the effectiveness of our approach, we implemented 
our FedEntropy framework using PyTorch. All  the experimental results 
were obtained on a Ubuntu workstation with 
Intel i9-10900k CPU, 32GB RAM, and NVIDIA GeForce RTX 3080 GPU.

\subsection{Experimental Settings}\label{4.1}

\textbf{Baseline Methods.}
We compared FedEntropy with four FL methods including the classic one and three state-of-the-arts, 
i.e., FedAvg \cite{mcmahan2017communication}, FedProx \cite{li2020federated}, SCAFFOLD \cite{karimireddy2019scaffold} and Moon \cite{li2021model}. Note that,
by default, we used FedAvg as the FL optimizer in FedEntropy.


\noindent\textbf{Dataset and DNN Settings.}
To facilitate  the performance comparison, we evaluated the FL methods on 
three well-known benchmark
datasets, i.e., 
CIFAR-10, CIFAR-100 \cite{krizhevsky2009learning} and CINIC-10 \cite{DBLP:journals/corr/abs-1810-03505}, where 
the images of CINIC-10 come from ImageNet \cite{DBLP:journals/corr/ChrabaszczLH17} and CIFAR-10. 
Note that the CIFAR-100 dataset has two kinds of sample labels, i.e., fine-grained labels (with 100 classes) and coarse-grained labels (with 20 superclasses). 
To better evaluate
the performance of different FL methods, we adopted the 
coarse-grained labels for classification. 
Similar to dataset distribution 
settings in 
\cite{wang2020optimizing,DBLP:conf/iclr/AcarZNMWS21,DBLP:conf/icml/YurochkinAGGHK19}, 
we adopted three kinds of data distributions to reflect the
data heterogeneity:
i) {\it case 1} where data on each device belong to the same single label;
ii) {\it case 2} where data on each device belong to two labels evenly;
and iii) {\it case 3} where   data   follow the 
Dirichlet distribution ($\beta=0.1$ by default). Note that 
we can tune the value of  $\beta$ to control the heterogeneity 
of non-IID data among devices, where a smaller $\beta$ indicates higher data heterogeneity. For all the three datasets, we used the DNN models with the same 
architecture. To be specific, each  DNN model here has   two 5$\times$5 convolution layers. The first layer has 6 output channels and the second layer has 16 output channels, where each layer is followed by a 2$\times$2 max pooling operation. 
Please refer to Appendix \ref{settings} for more details about the settings for datasets and DNNs.

\begin{table*}[th]
\vspace{-0.15in}
  \caption{Test accuracy comparison  for  different non-IID scenarios on  three datasets.}
  \footnotesize
  \label{t1}
  \setlength{\tabcolsep}{1mm}
  \centering
  \begin{tabular}{c|c|ccccc}
    \hline
    \multirow{2}{*} {Dataset} & \multirow{2}{*}{\makecell[c]{Heterogeneity \\ Settings}}  & \multicolumn {5}{c} {Test Accuracy($\%$)} \\ \cline{3-7}
     & & FedAvg &FedProx &SCAFFOLD &Moon &Ours\\
    \hline\hline
    \multirow {3}{*}{CIFAR-10} & case 1 &27.08$\pm$0.79 &27.59$\pm$0.55 &22.30$\pm$1.85 &22.10$\pm$1.30 &\textbf{32.52}$\pm$1.05\\
    &case 2  &48.90$\pm$0.47 &49.22$\pm$0.25 &48.73$\pm$1.62 &46.86$\pm$0.39 &\textbf{51.39}$\pm$1.18\\
    &case 3   &48.31$\pm$0.83 &47.77$\pm$0.78 &50.49$\pm$1.21 &47.38$\pm$0.53 &\textbf{51.00}$\pm$0.23\\
    \hline
    \multirow {3}{*}{CIFAR-100} & case 1  &15.39$\pm$0.86 &15.13$\pm$0.96 &14.49$\pm$0.73 &11.86$\pm$0.49 &\textbf{18.47}$\pm$0.75\\
    &case 2   &27.62$\pm$0.44 &27.25$\pm$0.14 &28.42$\pm$0.22 &27.11$\pm$0.0.58 &\textbf{29.24}$\pm$0.70\\
    &case 3   &28.71$\pm$0.22 &28.47$\pm$0.49 &\textbf{30.15}$\pm$0.50 &28.13$\pm$0.21 &29.21$\pm$0.25\\
    \hline
    \multirow {3}{*}{CINIC-10} & case 1  &23.23$\pm$1.87 &23.54$\pm$1.07 &20.27$\pm$0.90 &20.34$\pm$1.76 &\textbf{29.34}$\pm$1.48\\
    &case 2   &36.09$\pm$0.85 &36.63$\pm$0.97 &35.86$\pm$1.68 &36.12$\pm$0.92 &\textbf{37.65}$\pm$0.45\\
    &case 3   &34.11$\pm$1.39 &34.35$\pm$0.94 &\textbf{37.52}$\pm$0.43 &33.90$\pm$0.57 &35.84$\pm$0.70\\
    \hline
  \end{tabular}
\end{table*}

\begin{table*}[th]
\vspace{-0.15in}
  \caption{Communication overhead comparison for different non-IID scenarios on CIFAR-10.}
  \label{t2}
   \footnotesize
  \centering
  \begin{tabular}{c | c c c}
    \hline
    \multirow{2}{*}{FL Method} &\multicolumn {3}{c}{Communication Rounds}\\ \cline{2-4}
    &{case 1(acc = 30\%)} &{case 2 (acc = 50\%)} &{case 3 (acc = 50\%)} \\
    \hline\hline
    {FedAvg} &591.33$\pm$25.32 &316.00$\pm$37.59 &359.67$\pm$239.12 \\
    {FedProx} &569.33$\pm$122.50 &248.00$\pm$22.61 &279.33$\pm$48.09 \\
    {SCAFFOLD} &793.00$\pm$159.42 &339.67$\pm$24.95  &291.33$\pm$15.89 \\
    {Moon} &853.67+67.83 &297.33$\pm$35.16  &310.00$\pm$45.90 \\
    {Ours} &\textbf{340.00}$\pm$52.20  &\textbf{193.33}$\pm$13.50  &\textbf{223.67}$\pm$69.06 \\
    \hline
  \end{tabular}
\end{table*}

\noindent\textbf{Hyperparameters.} 
For  FedEntropy, we set the threshold $\epsilon$ to  0.8. 
For FedProx, we set the  hyperparameter $\mu$ to 0.01, which controls  the proximal term weight. 
For SCAFFOLD, we set the global step-size $\eta_g$ to 1. 
Similar to the settings in \cite{li2021model}, for Moon
 we set the hyperparameters $\mu$ to 0.1 and  temperature $\tau$ to 0.5.
For all FL methods, we set the number of clients (i.e., $N$) to  100, and the 
number of communication rounds (i.e., $T$) to 1000. 
To mimic   real-world scenarios with limited communication 
resources, we set the ratio of participant devices in each 
FL training round to  10\% (i.e., $|S_t|$ = 10). 
 For each device, the local optimizer is $SGD$ with a learning rate of 0.01 and a momentum of 0.5. For local training, we set the batch size to 50 and the number of local training epochs (i.e., $E$) to  5.


\subsection{Performance Comparison}\label{4.2}

\begin{figure*}[t]
\vspace{-0.2in}
\centering
\subfloat[CIFAR-10 with case 1]{\includegraphics[width=.25\linewidth]{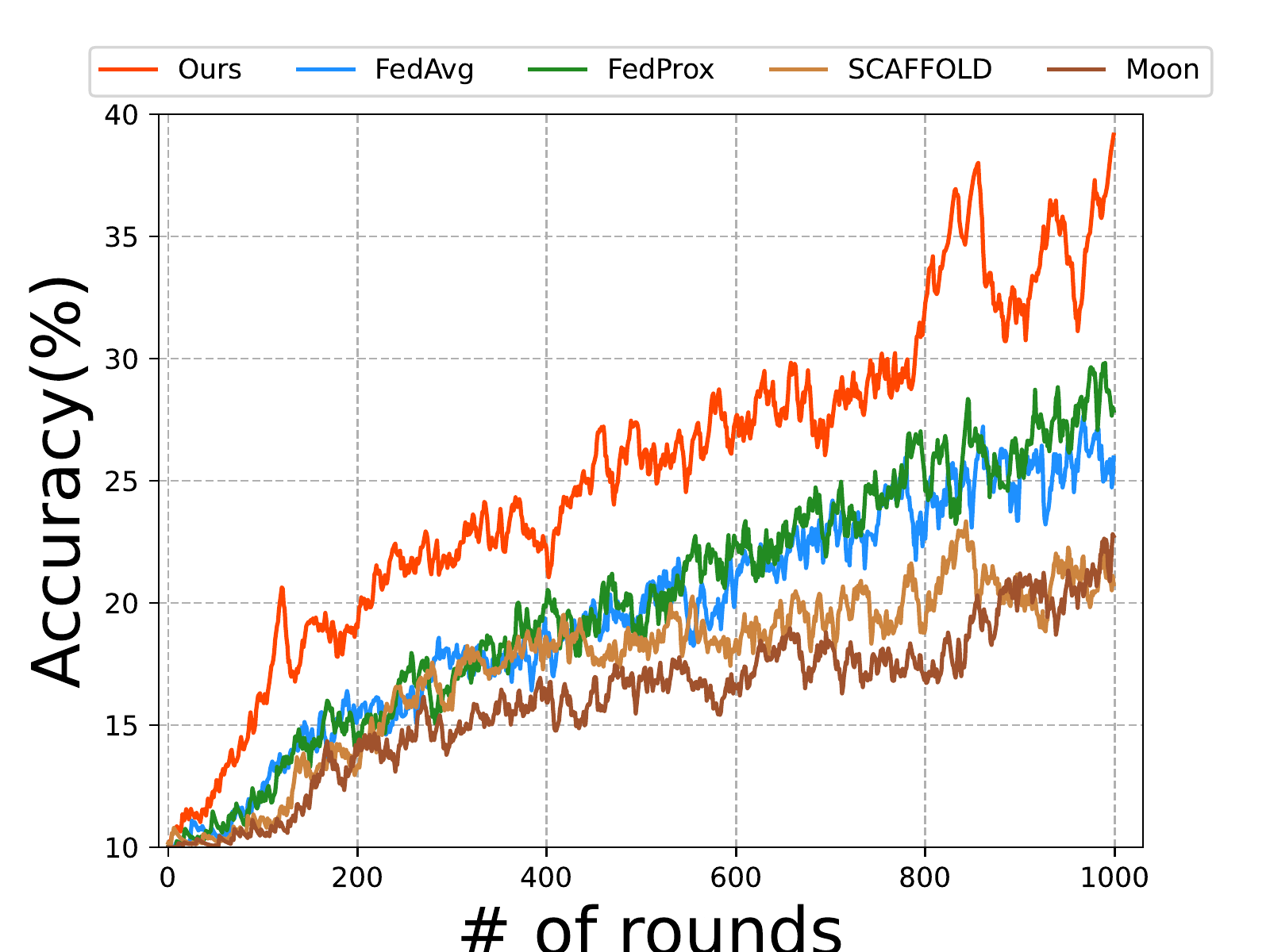}}
\subfloat[CIFAR-10 with case 2]{\includegraphics[width=.25\linewidth]{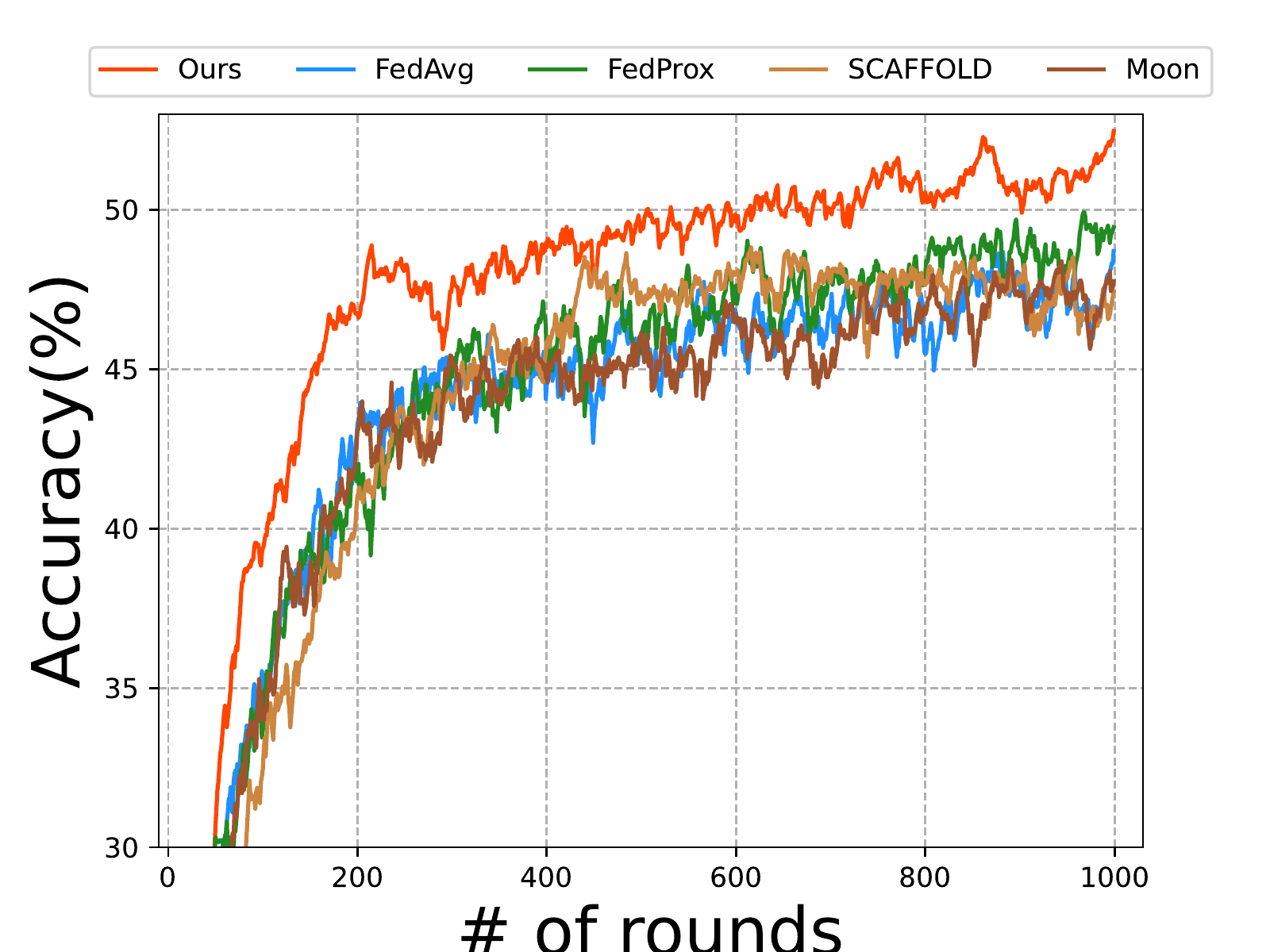}}
\subfloat[CIFAR-10 with case 3]{\includegraphics[width=.25\linewidth]{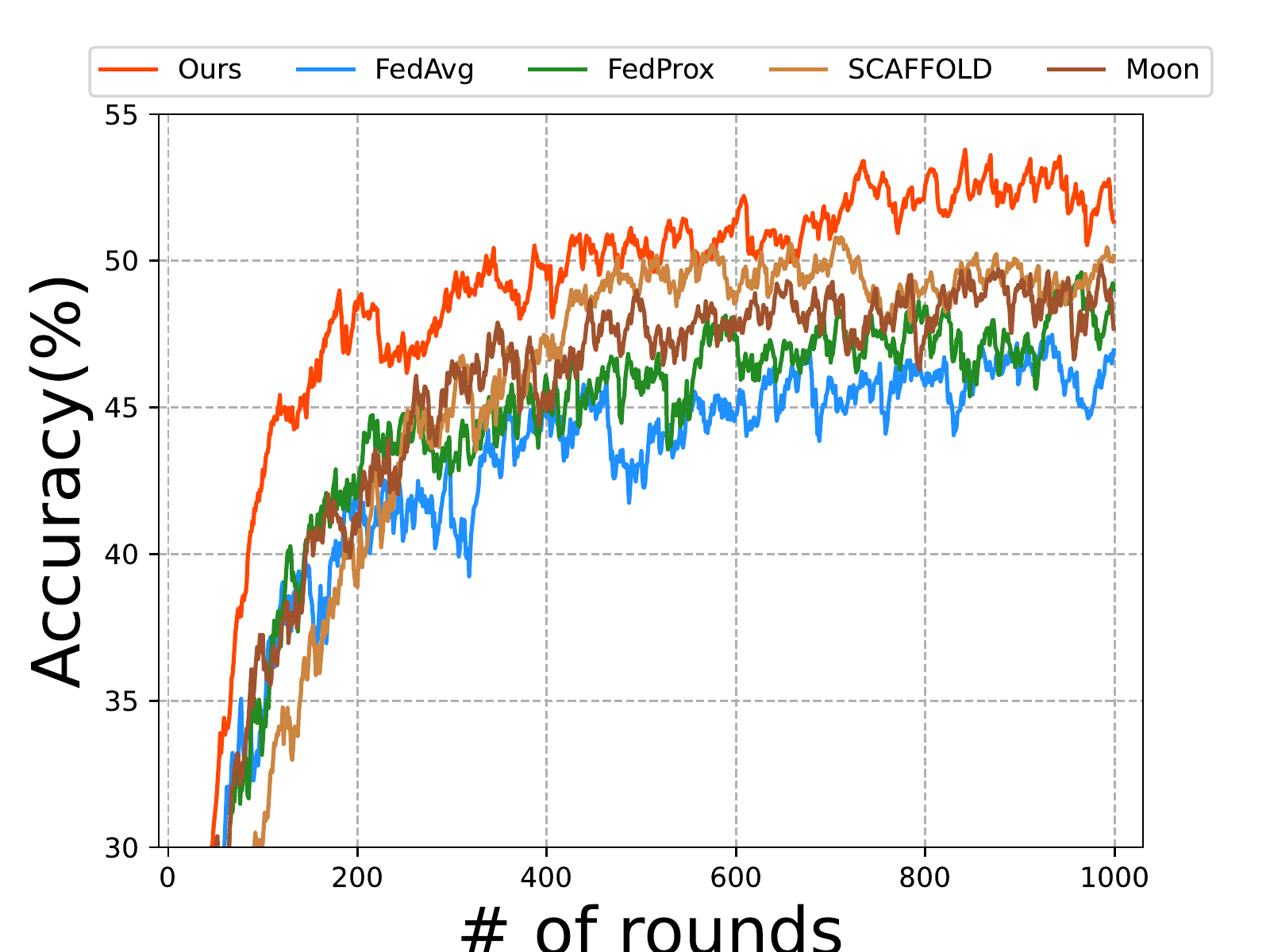}}\\
\subfloat[CIFAR-100 with case 1]{\includegraphics[width=.25\linewidth]{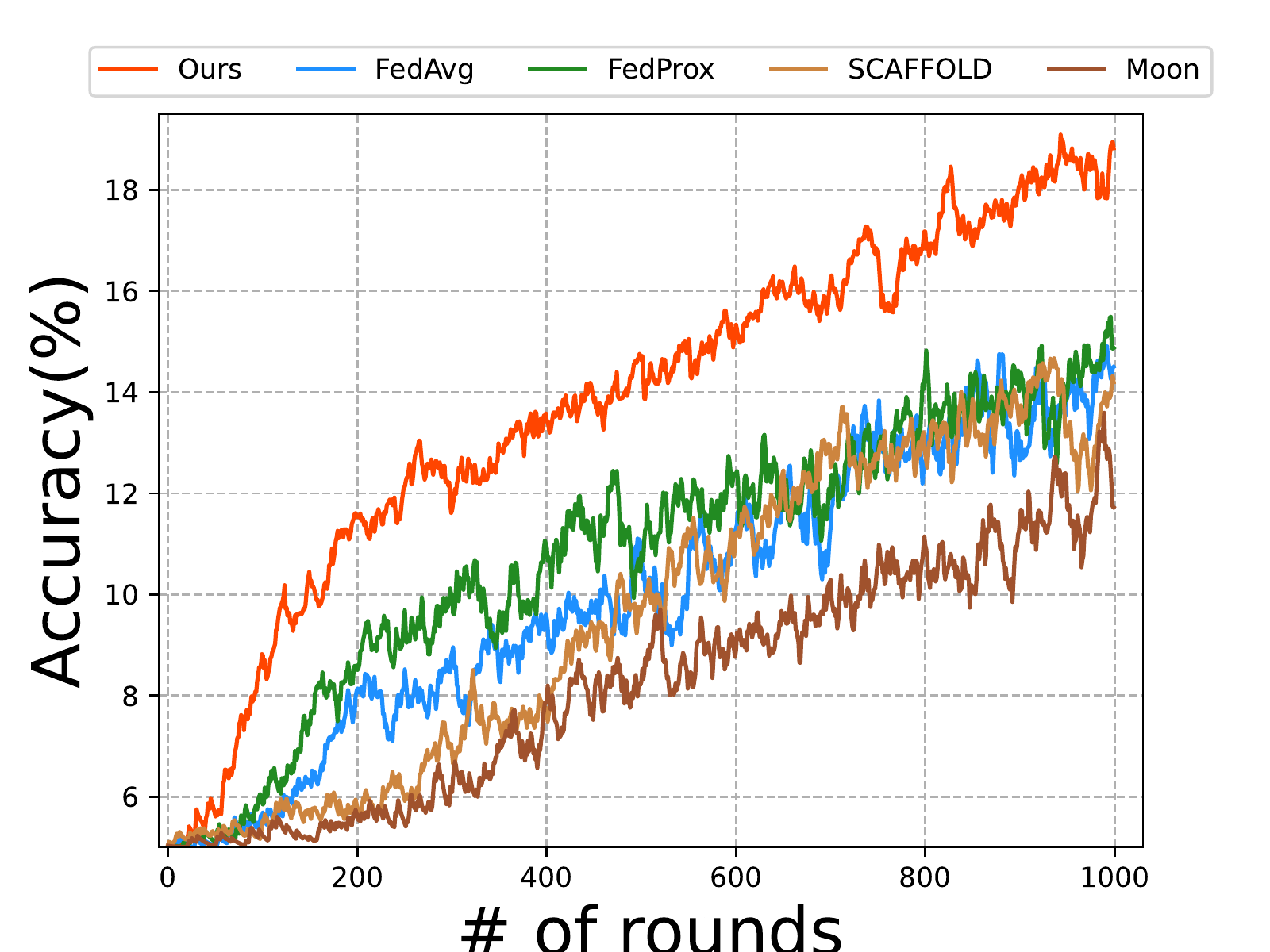}}
\subfloat[CIFAR-100 with case 2]{\includegraphics[width=.25\linewidth]{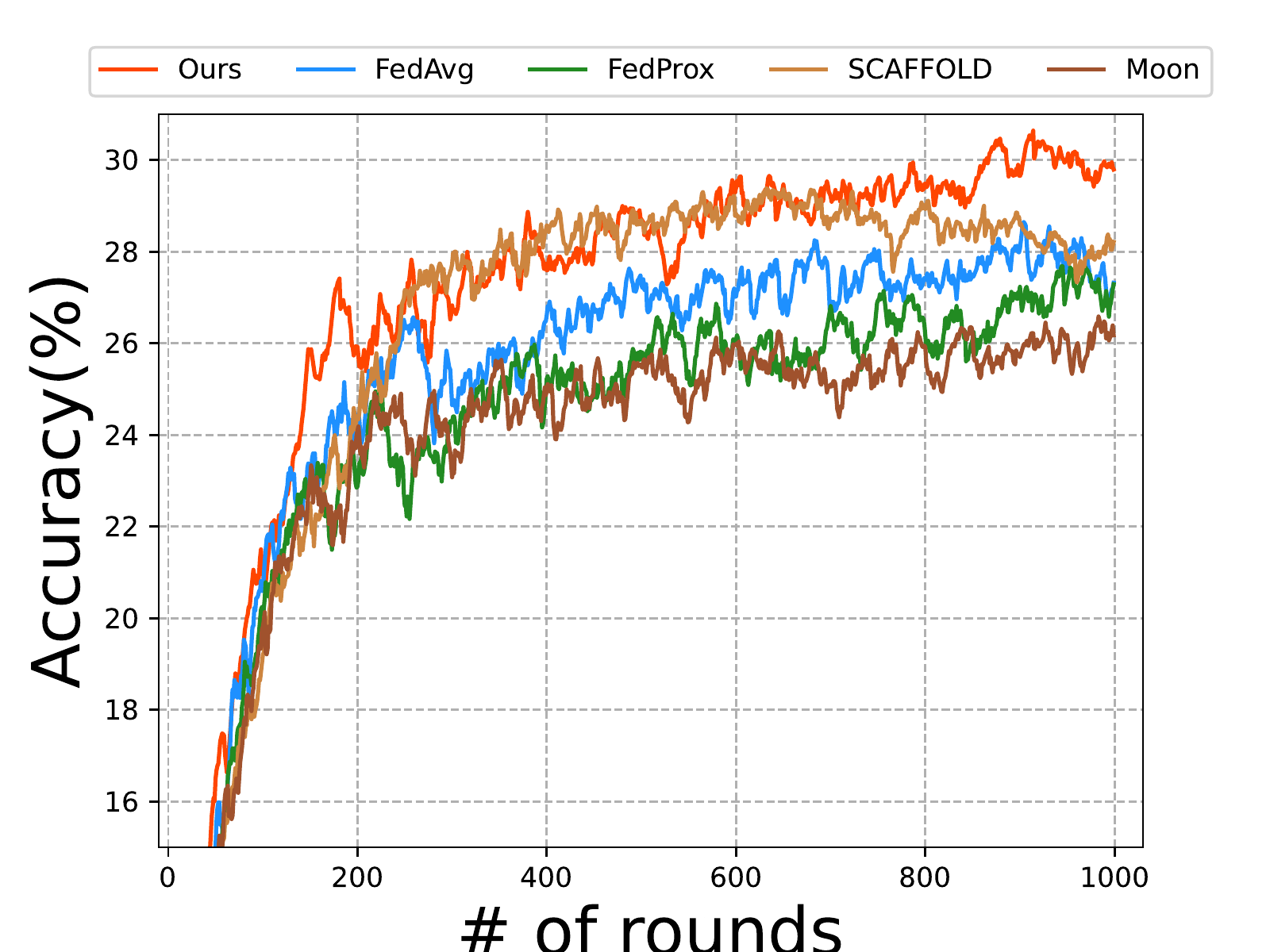}}
\subfloat[CIFAR-100 with case 3]{\includegraphics[width=.25\linewidth]{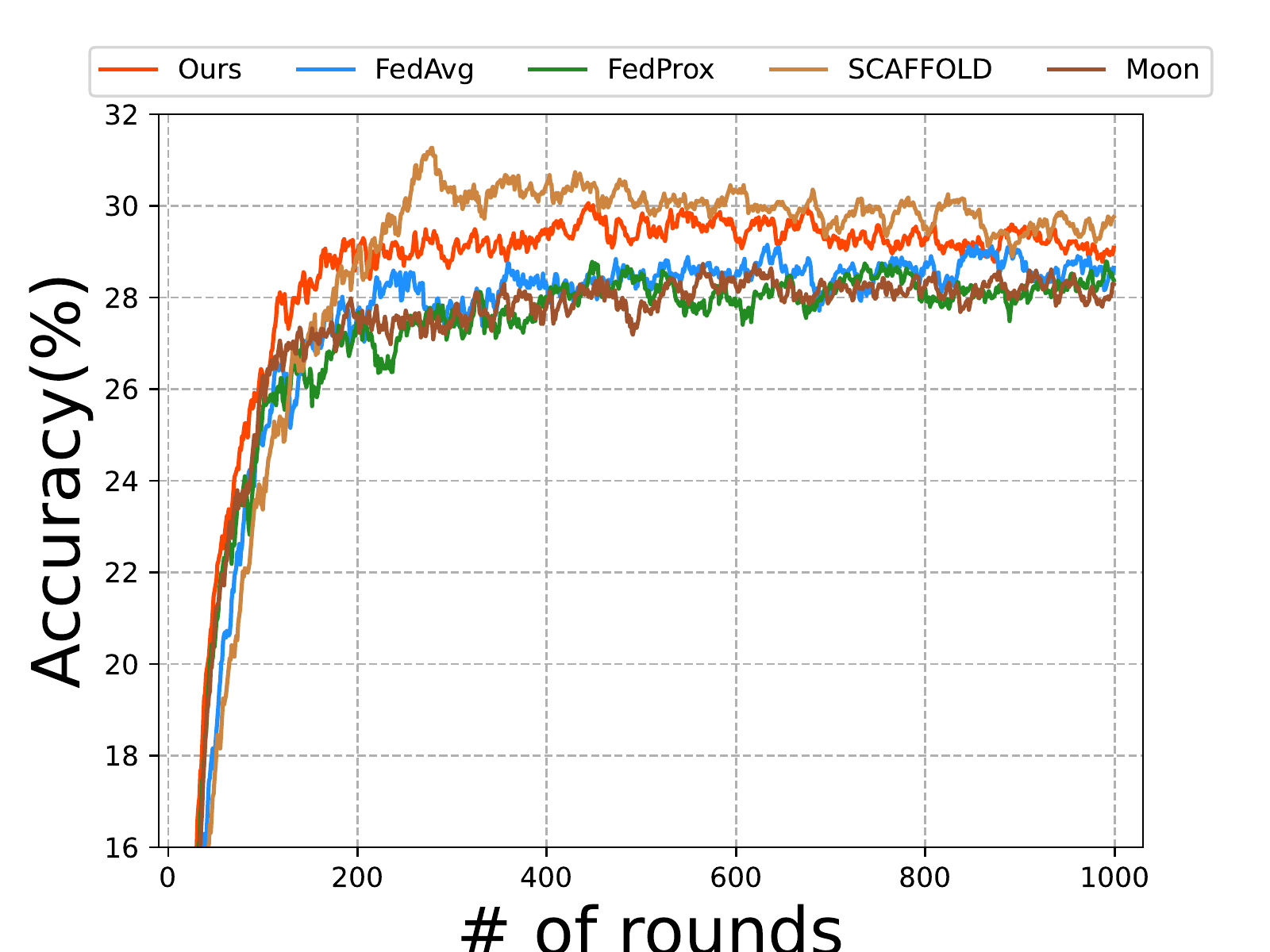}}\\
\subfloat[CINIC-10 with case 1]{\includegraphics[width=.25\linewidth]{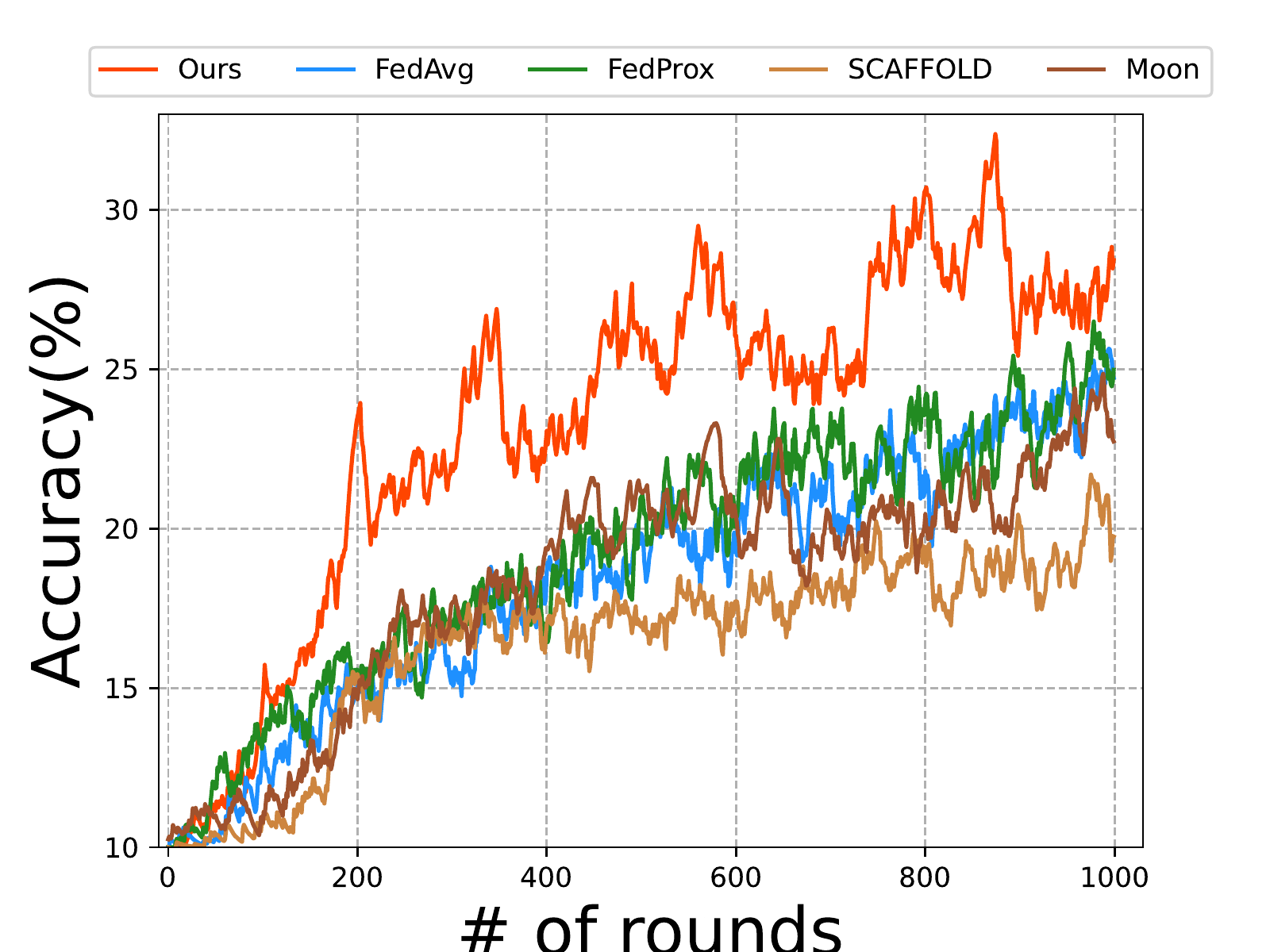}}
\subfloat[CINIC-10 with case 2]{\includegraphics[width=.25\linewidth]{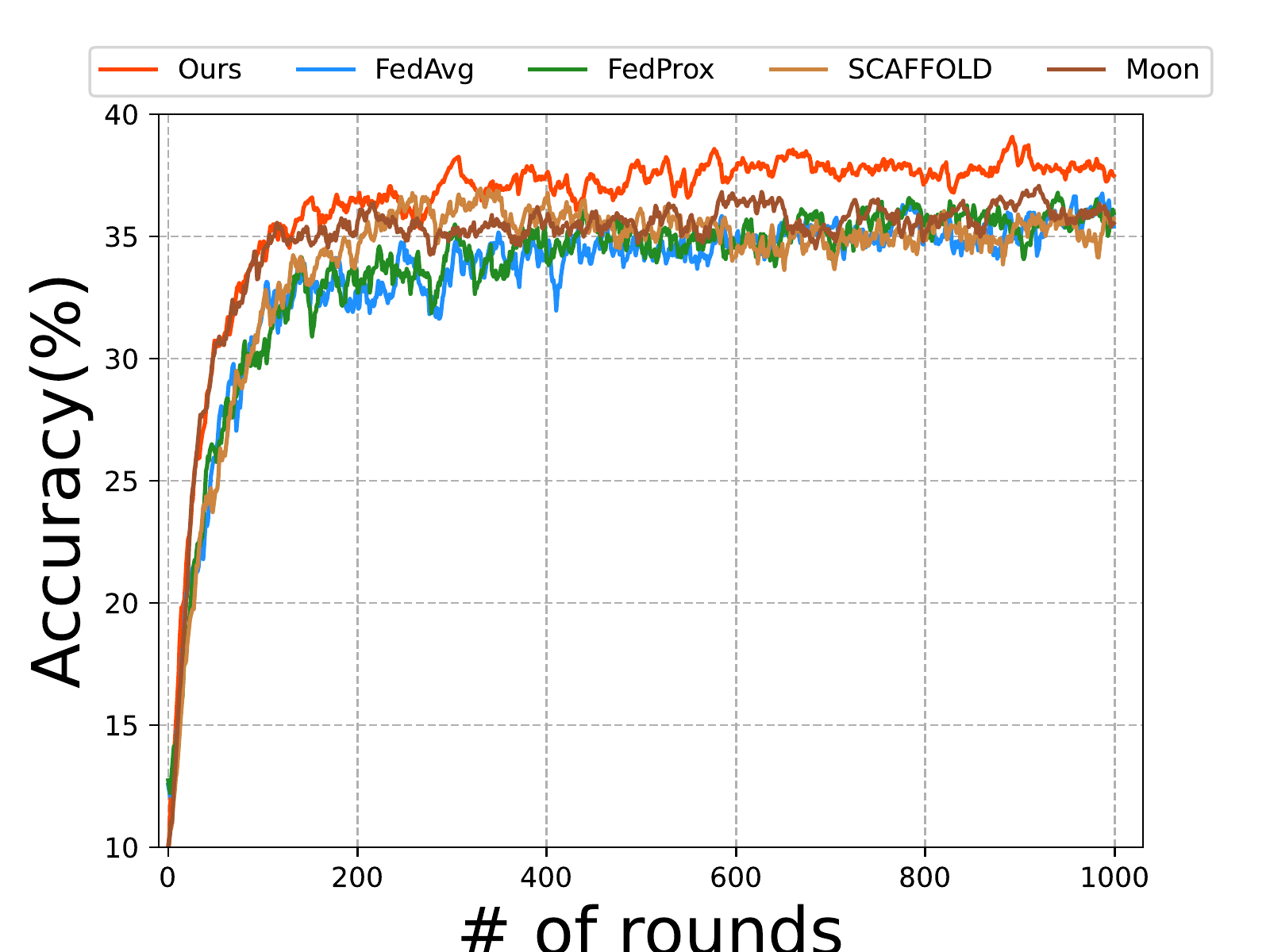}}
\subfloat[CINIC-10 with case 3]{\includegraphics[width=.25\linewidth]{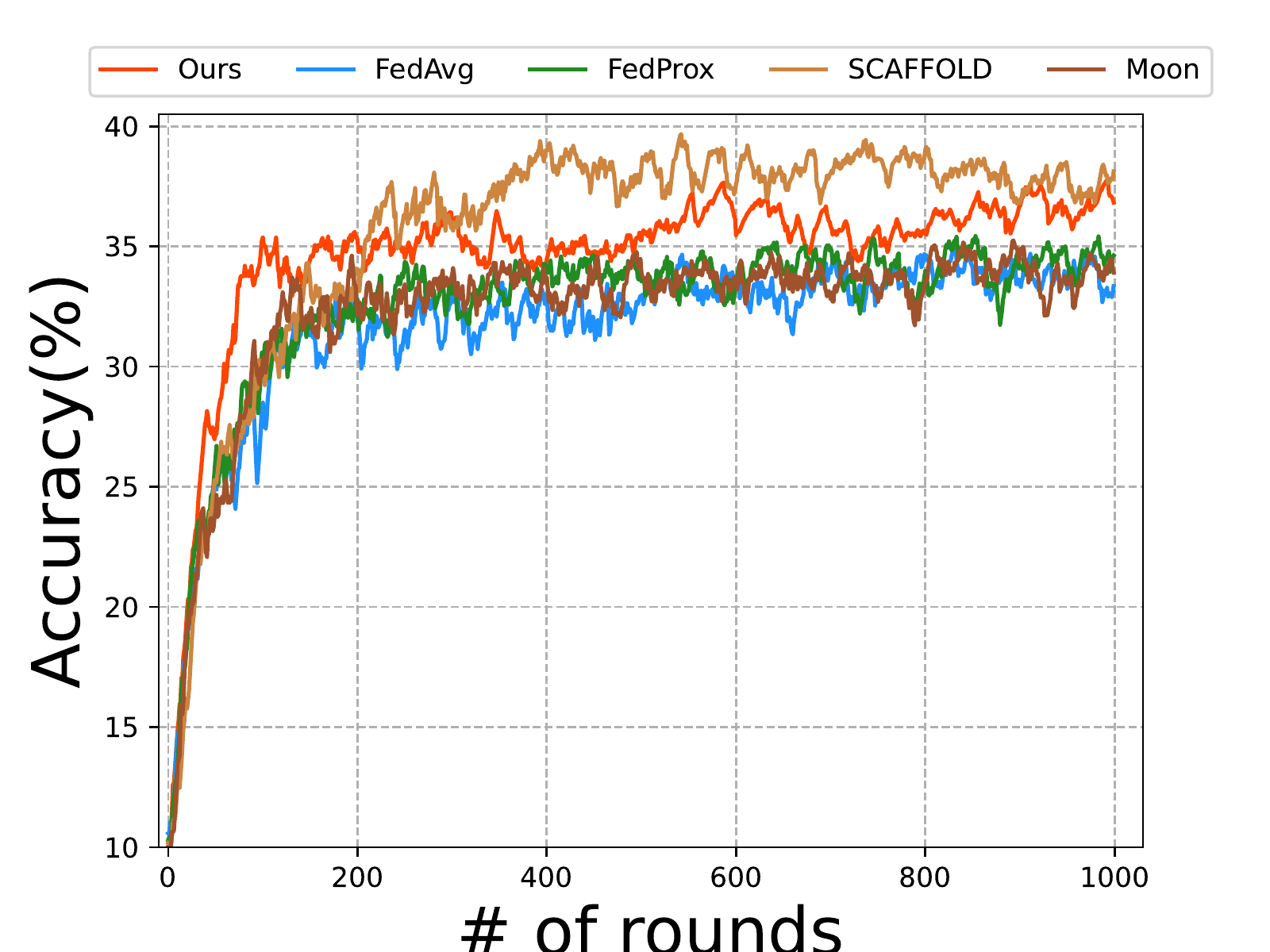}}
\caption{Convergence  comparison between different FL methods for 
different non-IID scenarios.}
\label{f2}
\end{figure*}

\textbf{Test Accuracy.}
Table \ref{t1} presents the performance comparison results between our FedEntropy and the four baseline methods in terms of test accuracy. To enable a fair comparison, all the experiments were repeated  3 times with different random seeds, and we calculated
the average results of the last ten rounds. 
From Table \ref{t1}, we can find
that under different data heterogeneity settings, FedEntropy outperforms other FL methods
with a considerable improvement.
Taking the case 1 of  CIFAR-10 as an example, the test accuracy of our FedEntropy is 5.44\%, 4.93\%, 10.22\%, and 10.42\% higher than FedAvg, FedProx, SCAFFOLD, and Moon, respectively. 
Figure \ref{f2} presents the  accuracy trends of all the five
studied FL methods on the three  datasets with  different data heterogeneity settings. 
 From  Figure \ref{f2}, we can find that in most cases FedEntropy can
 not only achieve the highest accuracy, but also converge faster than the other FL methods for a given accuracy.

\begin{table*}[th]
\vspace{-0.1in}
  \caption{Impact of FL optimizers on FedEntropy (on dataset CIFAR-10).}
  \label{t3}
  \setlength{\tabcolsep}{1mm}
  \centering
  \begin{tabular}{c | c c c}
    \hline
  \multirow{2}{*}{Combination}  & \multicolumn {3}{c}{Test Accuracy($\%$)}\\ 
    \cline{2-4}
    &case 1 &case 2 &case 3\\
    \hline\hline
    {FedAvg + FedEntropy} &\textbf{40.74}$\pm$0.88 &\textbf{54.21}$\pm$0.12 &\textbf{53.96}$\pm$1.55\\
    {FedProx + FedEntropy} &40.44$\pm$1.81 &53.78$\pm$0.87 &53.37$\pm$0.56\\
    {SCAFFOLD + FedEntropy} &30.20$\pm$1.14 &52.28$\pm$0.89 &53.47$\pm$0.21\\
    {Moon + FedEntropy} &35.97$\pm$2.78 &53.91$\pm$0.87 &52.92$\pm$0.83\\
    \hline
  \end{tabular}
\end{table*}

\noindent\textbf{Communication Overhead.}
Considering that the downlink communication overhead of all FL methods is consistent,  here 
we  only consider the  communication overhead of uploading local models. 
For FedAvg, FedProx and Moon, the communication overhead per round is only for uploading the models of  selected devices.
However,
for SCAFFOLD, in each round it additionally requires uploading the local control variables of  selected devices, whose  sizes are the same as 
the ones of local models. 
In other words, the communication cost of SCAFFOLD is twice that of traditional FL methods.
In the first stage of device selection, 
FedEntropy firstly  uploads the soft labels of selected
devices to the cloud. Since  the data volume of  soft labels is much 
smaller than one of the local models, such communication overhead is negligible.  For the second stage, 
under maximum entropy judgment,  the communication overhead
 of FedEntropy per round is at most the same as the one of  FedAvg. 
Table \ref{t2} presents
the communication overhead  required by the 
five FL methods to achieve the same accuracy for different non-IID scenarios of CIFAR-10.
For the three different non-IID scenarios, we set the  expected accuracy to  30\%, 50\%, and 50\%, respectively. We can find that FedEntropy has the least number of communication rounds and the least communication overhead in all scenarios. Note that we can observe the similar trends for
the other two datasets. 

\noindent\textbf{Synergy between FedEntropy and  other FL methods.}
Table \ref{t3} presents the performance of FedEntropy using FedAvg, FedProx, SCAFFOLD, and Moon as optimizers. 
Compared  with the results in 
Table \ref{t1}, we can 
find that the classification
performance of all the  four FL methods under the help of our 
maximum entropy-based dynamic device grouping  is improved significantly.  
In other words, our dynamic device grouping  can be integrated into 
other FL methods to enhance their classification performance, 
showing the orthogonality of FedEntropy for other FL methods.
By simply applying our dynamic device grouping method on 
FedAvg, the results  in Table \ref{t1} and Table \ref{t3} 
show that the classification
performance of FedAvg+FedEntropy can even outperform SCAFFOLD.



\subsection{Ablation Study}\label{4.3}
\begin{figure*}[h]
\vspace{-0.2in}
\centering
\subfloat[Impact of FL optimizer.]{\includegraphics[width=.30\linewidth]{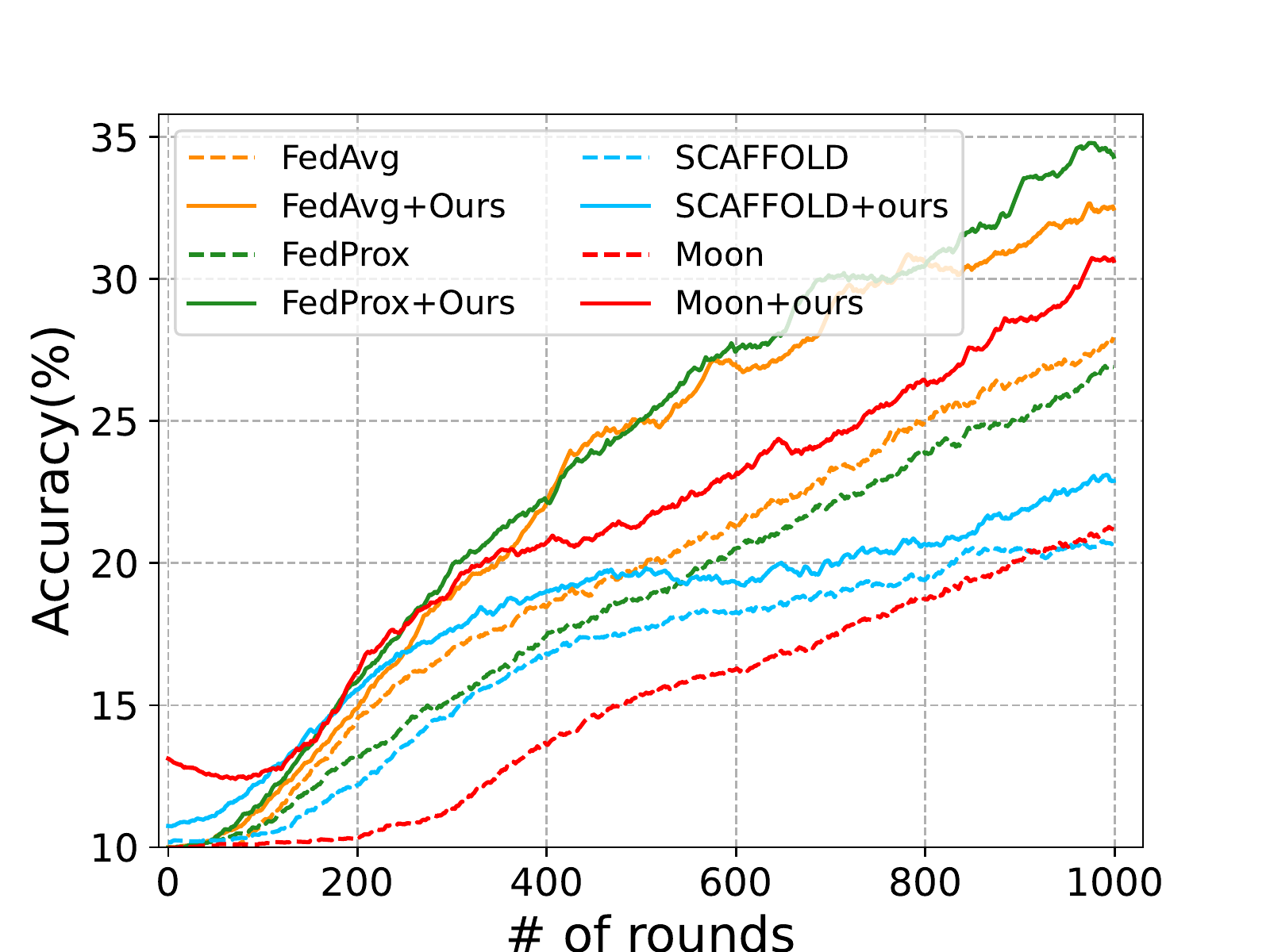}\label{f3.a}}\hspace{10mm}
\subfloat[Impact of different grouping strategies.]{\includegraphics[width=.30\linewidth]{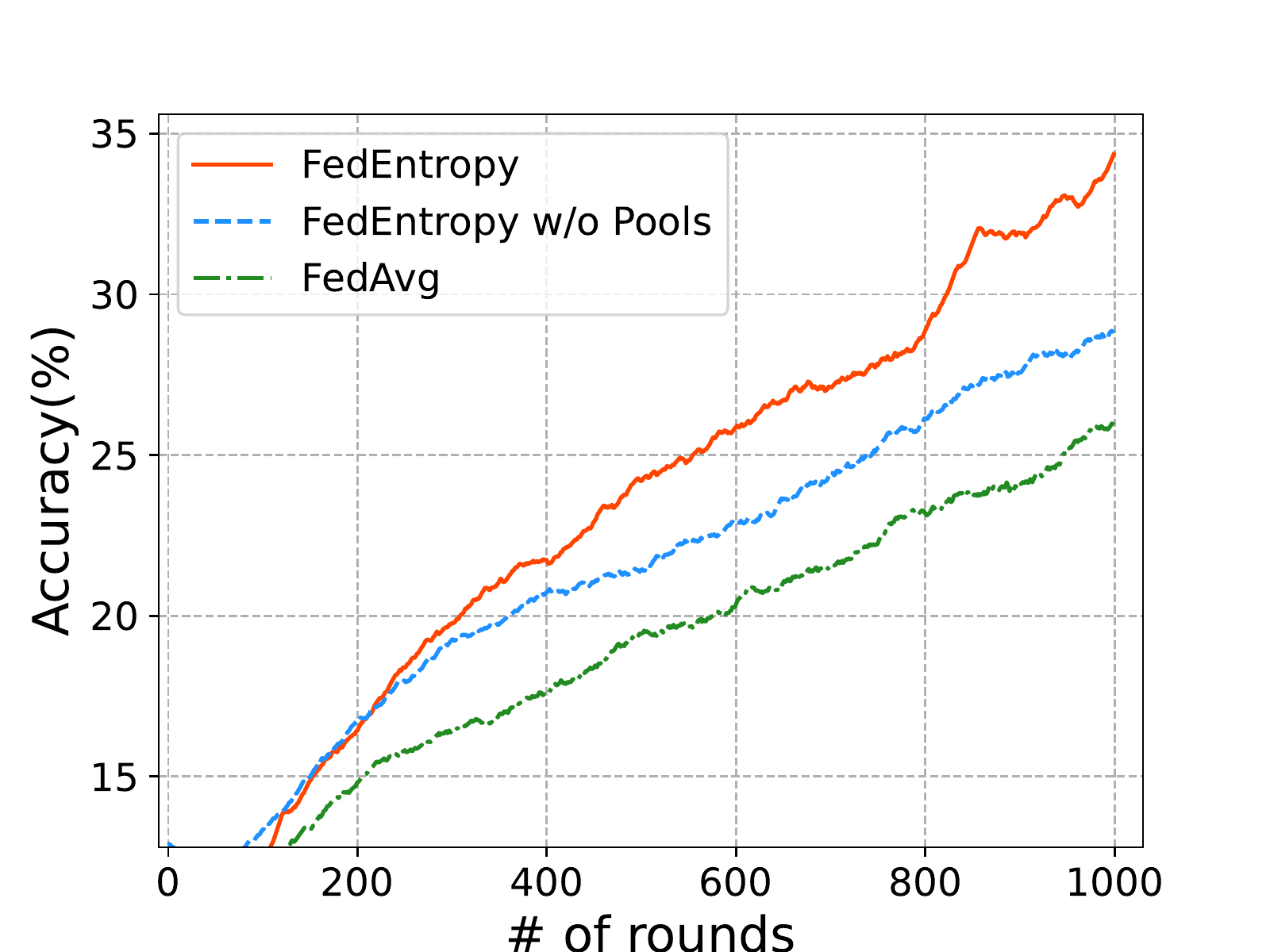}\label{f3.b}}
\caption{Impact of dynamic device grouping and its components.}
\end{figure*}

We conducted ablation studies to demonstrate the impact
of our proposed dynamic device grouping method used in 
FedEntropy and the impact of  components (maximum entropy judgment and 
positive/negative device pools)
in FedEntropy. 
Figure \ref{f3.a} shows the results on the case 1 of CIFAR-10. 
We can find that under the help of FedEntropy, 
both the accuracy and convergence  of  four FL baseline methods are 
dramatically improved. 
Please refer to Appendix \ref{Supplementary} for more 
results on other cases and datasets. 
Figure \ref{f3.b} investigates the impacts of key FedEntropy components.
Here, we consider three cases: i) FedEntropy, ii) FedEntropy without
positive/negative  device pools, and iii) FedAvg. 
Note that for the second case, we assume that all the selected devices are all put in the positive device pool. 
We can find that 
both two components on cloud introduced by our proposed
FedEntropy can indeed greatly increase the classification accuracy. 
Due to the space limitation, Figure \ref{f3.b} only presents the comparison results for the case 1 of CIFAR-10, the 
other cases with the same or different  datasets have the 
same trends as the one shown here.

\vspace{-0.15in}
\section{Conclusion}
\label{conclusion}

To accommodate the biased device data distributions in 
 non-IID scenarios, in this paper  we proposed a novel FL 
 framework named FedEntropy, which takes both the 
 data distributions of devices and 
 contributions of local models to global model aggregation
 into account. 
Based on our maximum entropy judgement heuristic,
we designed 
 a two-stage device grouping method to enable fine-grained 
 selection of more suitable local models to achieve better 
 aggregation quality. 
Comprehensive experimental results 
on various well-known benchmarks show that, compared with  state-of-the-art FL methods, 
FedEntropy  not only can achieve better FL performance in terms of both  classification accuracy and communication overhead, but also can be 
used to enhance  their classification performance.

\bibliographystyle{unsrt}
\bibliography{ijcai22}

\clearpage

\appendix

\section{Appendix}

\subsection{Dataset and Model Settings}
\label{settings}

Table \ref{datasets} details the  dataset settings used in the experiments.
Table \ref{model} presents the  settings for CNNs used in the experiments. 

\begin{table*}[th]
\caption{Dataset Settings}
\label{datasets}
\centering
\setlength{\tabcolsep}{1mm}{
\begin{tabular}{c c c c c }
\hline
Dataset  & Input Size  & Classes & Training Images & Test Images \\ \hline \hline
CIFAR-10 & 3$\times$32$\times$32 & 10      & 50000            & 10000     \\ 
CINIC-10   & 3$\times$32$\times$32 & 10      & 90000            & 90000      \\ 
CIFAR-100   & 3$\times$32$\times$32 & 20      & 50000            & 10000      \\ \hline
\end{tabular}} \label{dataset}
\end{table*}
\begin{table*}[th]
  \caption{CNN Model Settings}
  \label{model}
  \setlength{\tabcolsep}{1mm}
  \centering
  \begin{tabular}{c|c|c}
    \hline
    Layer Index &Type &Parameters \\
    \hline\hline
    1 &Convolutional &kernel: 5$\times$5, filter: 6 \\
    2 &Convolutional &kernel: 5$\times$5, filter: 16 \\
    3 &Fully Connected &unit: 120\\
    4 &Fully Connected &unit: 84\\
    5 &Fully Connected &unit: class numbers of dataset\\
    \hline
  \end{tabular}
\end{table*}

\subsection{Visualization of Data Heterogeneity}
\label{Data Heterogeneity}

\begin{figure*}[!ht]
\centering
\subfloat[CIFAR-10 (case 1)]{\includegraphics[width=.33\linewidth]{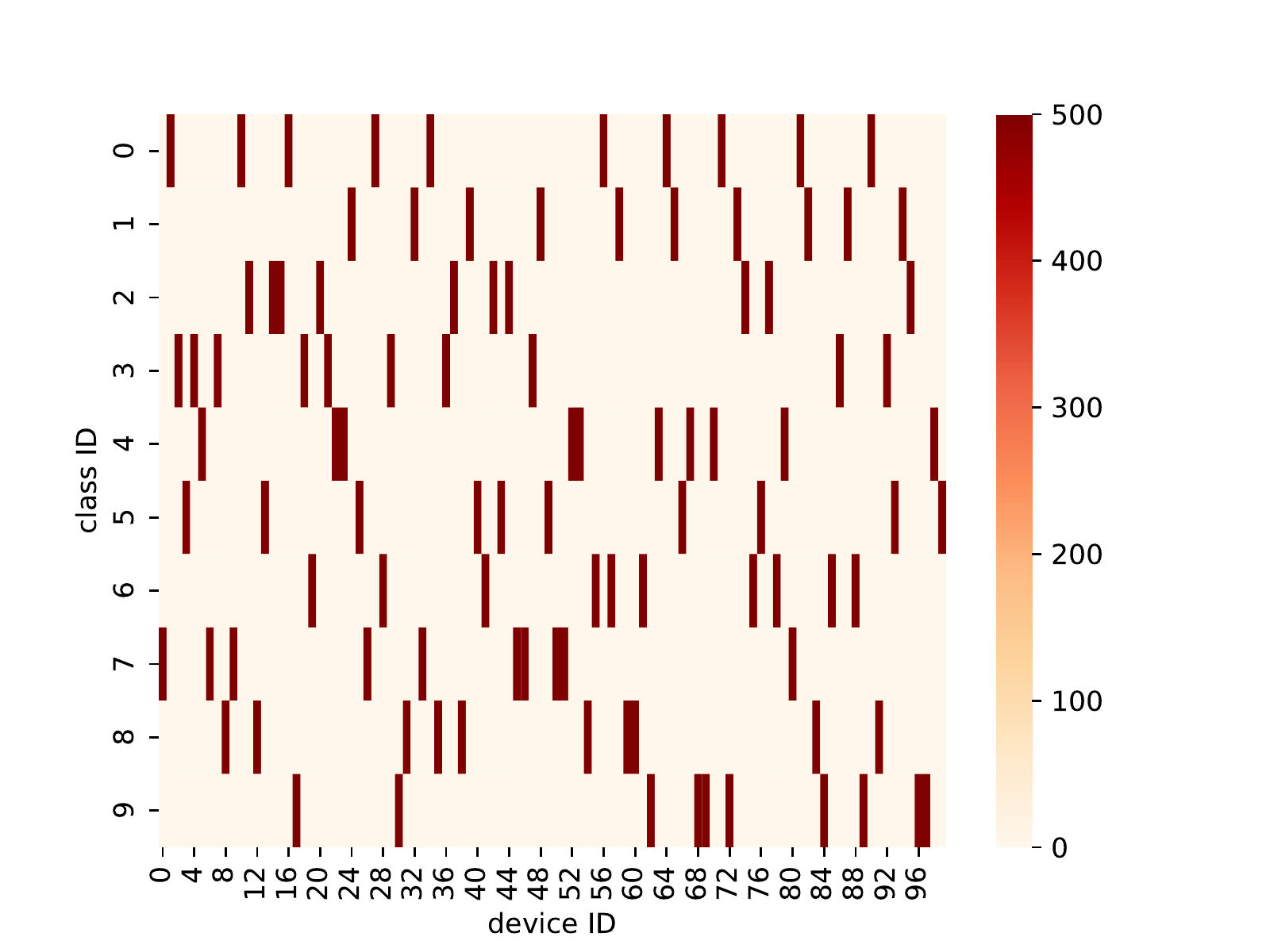}}
\subfloat[CIFAR-10 (case 2)]{\includegraphics[width=.33\linewidth]{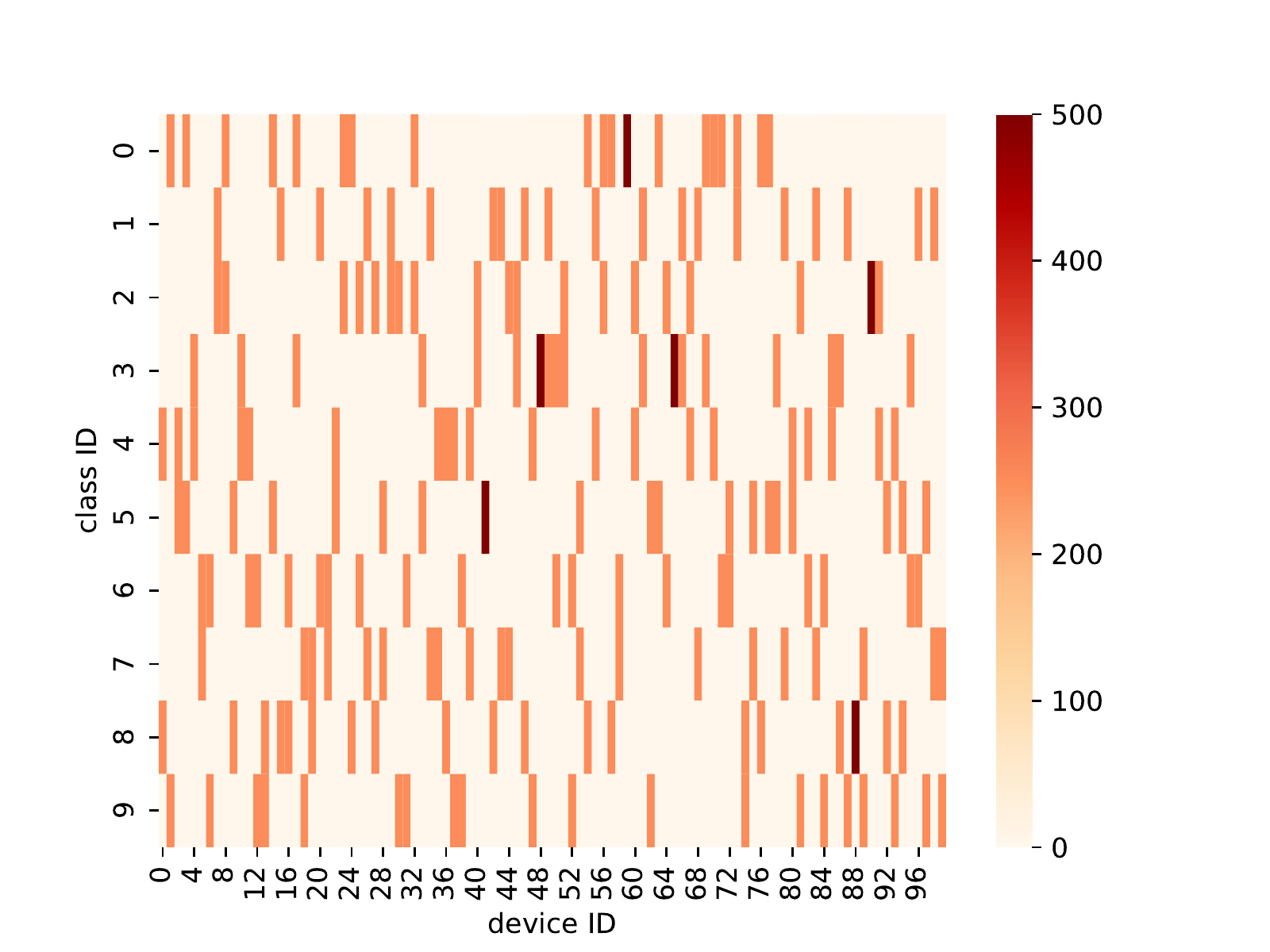}}
\subfloat[CIFAR-10 (case 3)]{\includegraphics[width=.33\linewidth]{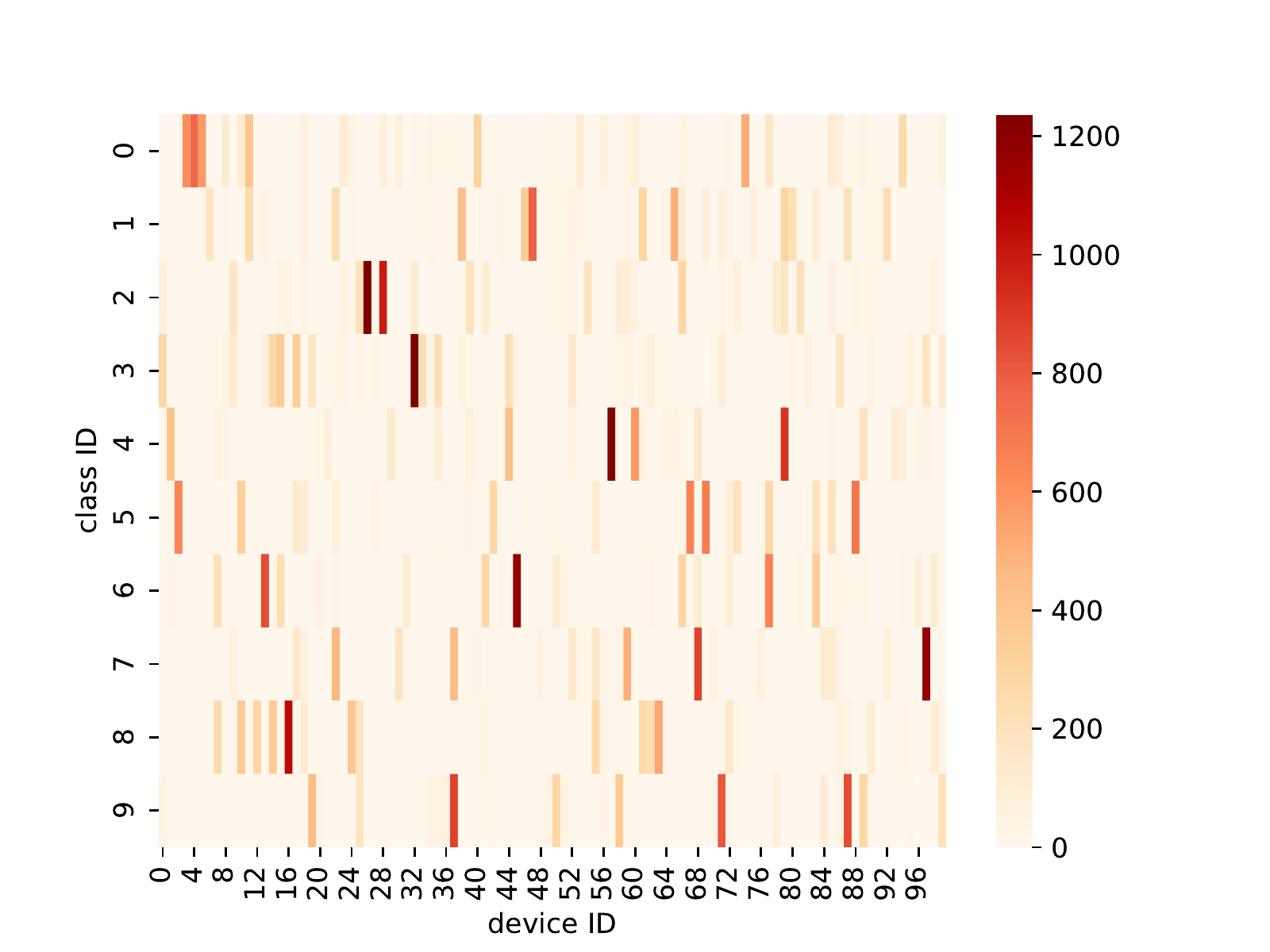}}\\

\subfloat[CIFAR-100 (case 1)]{\includegraphics[width=.33\linewidth]{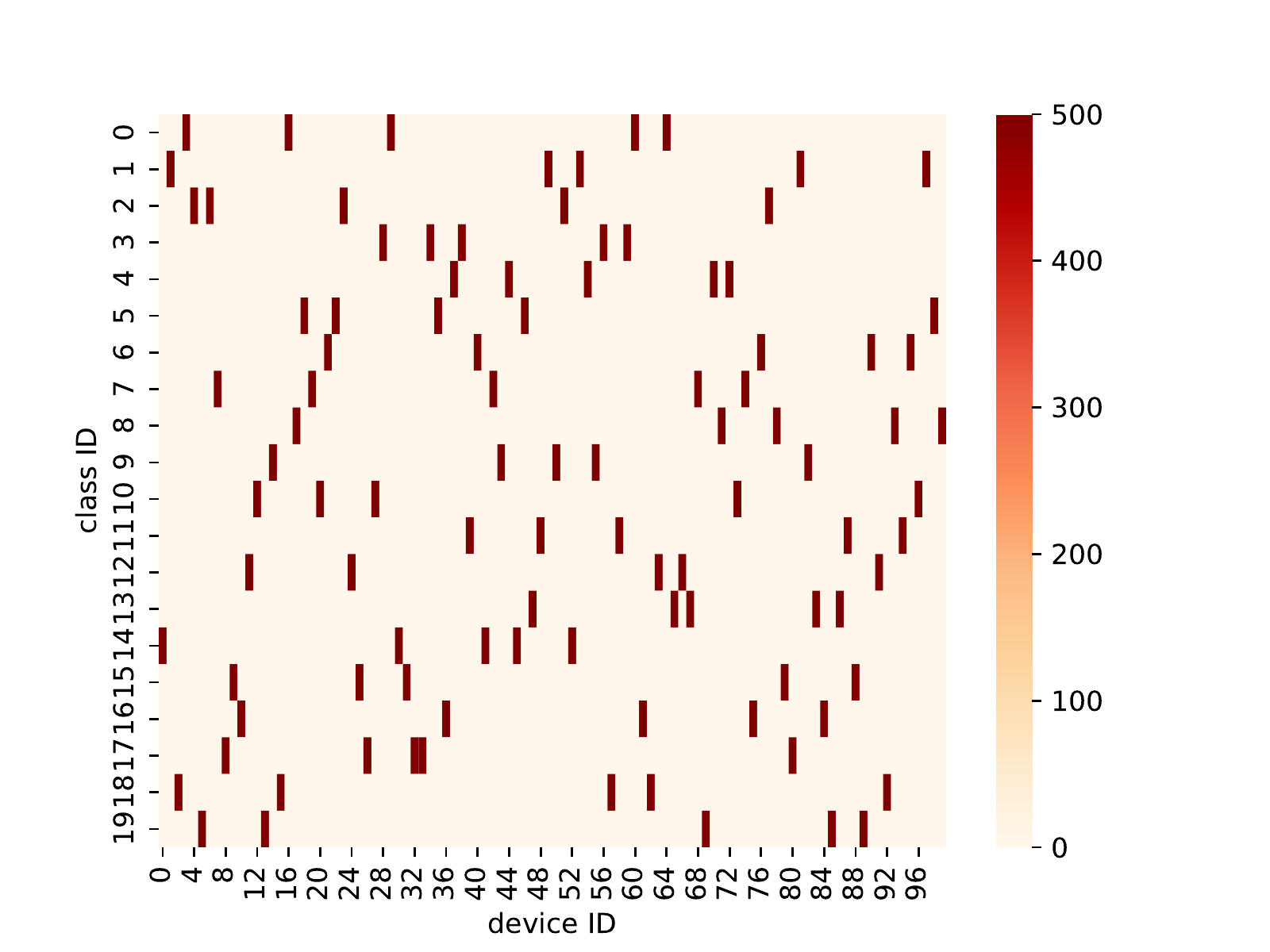}}
\subfloat[CIFAR-100 (case 2)]{\includegraphics[width=.33\linewidth]{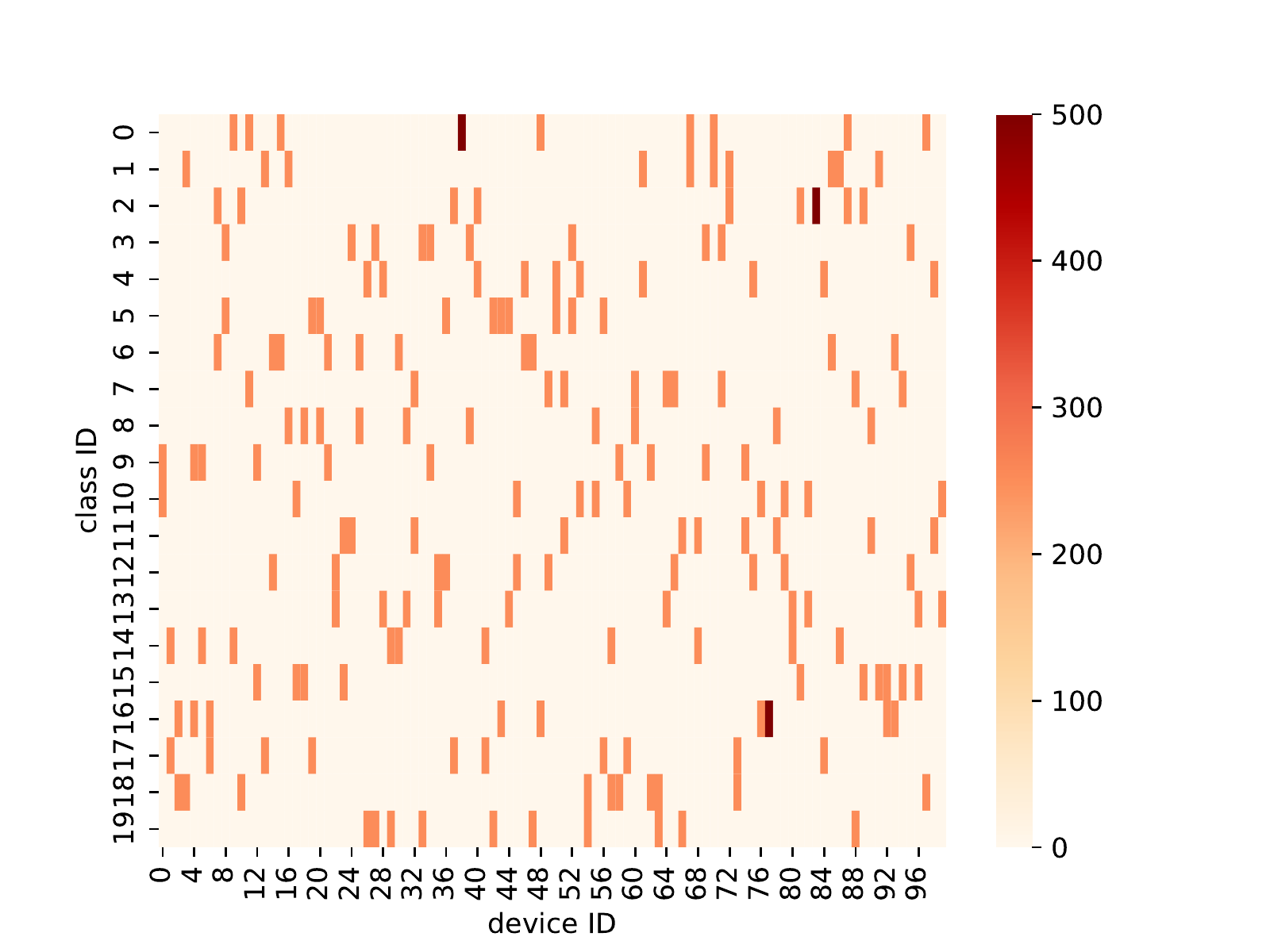}}
\subfloat[CIFAR-100 (case 3)]{\includegraphics[width=.33\linewidth]{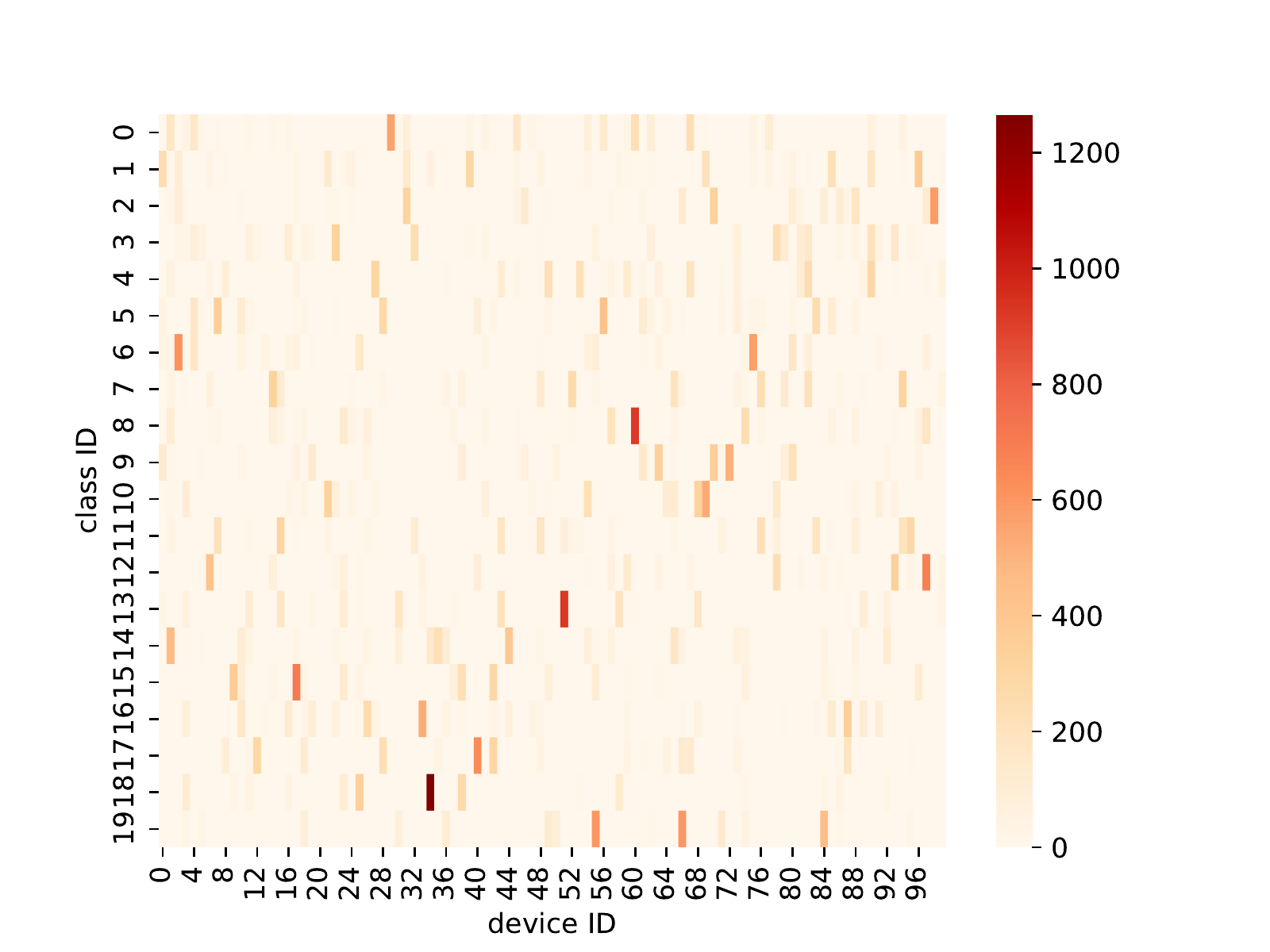}}

\caption{Data distributions in different cases.}
\label{f5}
\end{figure*}

Figure \ref{f5} shows the data distributions of clients generated by the  three heterogeneous data settings 
defined in  Section \ref{4.1}. In each subfigure, we  show the data distribution  of 100 clients. 
In Figure \ref{f5}, the data distribution of clients in different cases is distinctly different, and the degree of data heterogeneity increases from case 1 to case 3.


\subsection{Supplementary Experimental Results}
\label{Supplementary}
In this section, we provide more experiments to show the advantages of  FedEntropy, whose  code is available at \url{https://github.com/FedEntropy/FedEntropy}.

\noindent\textbf{Communication Overhead.}
For  given target test accuracy (i.e., 12\%, 28\%, 30\% correspond for  cases 1-3, respectively), Table \ref{t6} presents
the number of FL
communication rounds to reach this 
reach target. This information is a 
supplement to Table \ref{f2}.

\begin{figure*}[!ht]
\centering
\vspace{-0.15in}
\subfloat[CIFAR-10 with case 1]{\includegraphics[width=.33\linewidth]{images/Orthogonality-CIFAR10-case1.pdf}}
\subfloat[CIFAR-10 with case 2]{\includegraphics[width=.33\linewidth]{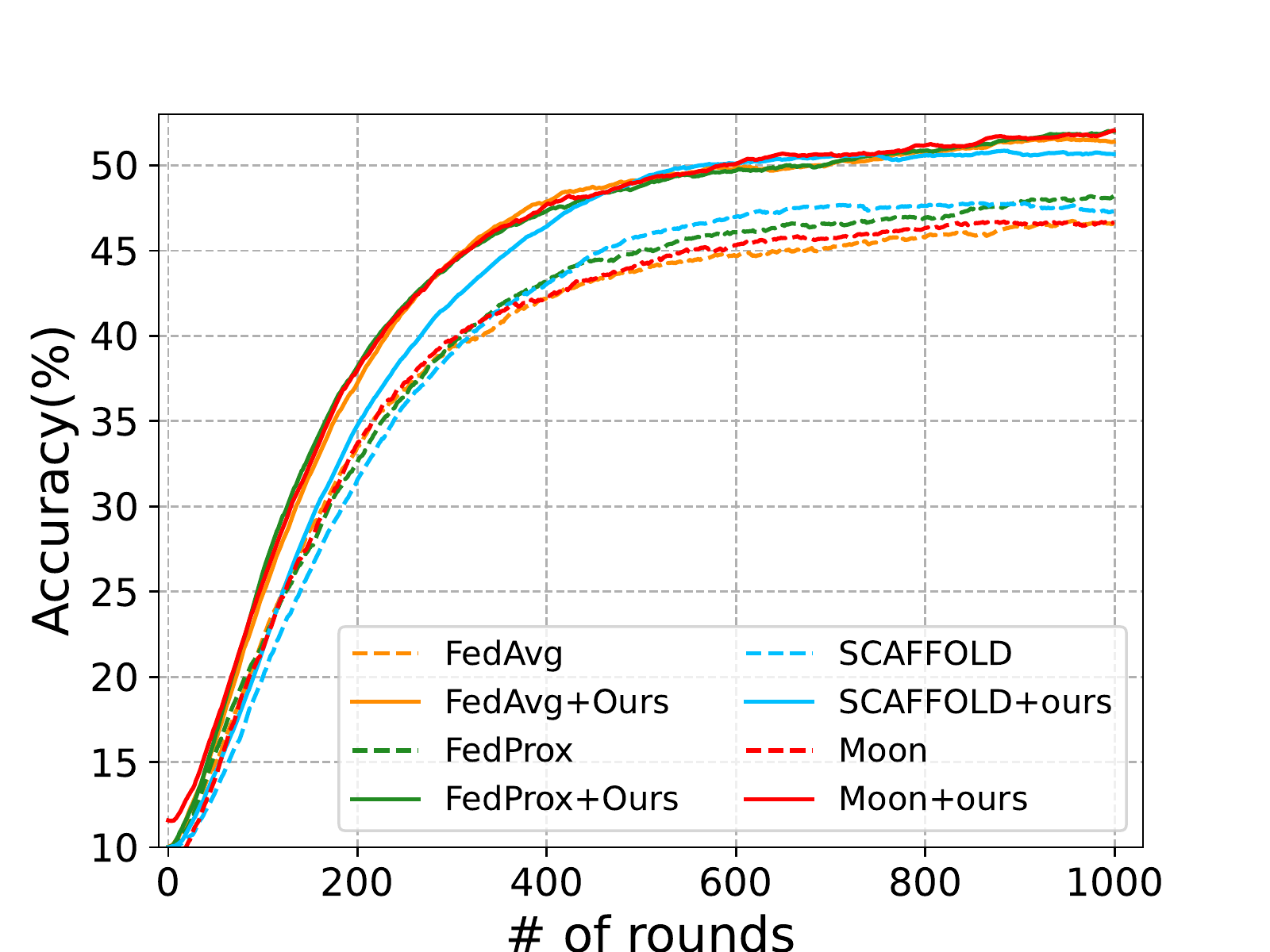}}
\subfloat[CIFAR-10 with case 3]{\includegraphics[width=.33\linewidth]{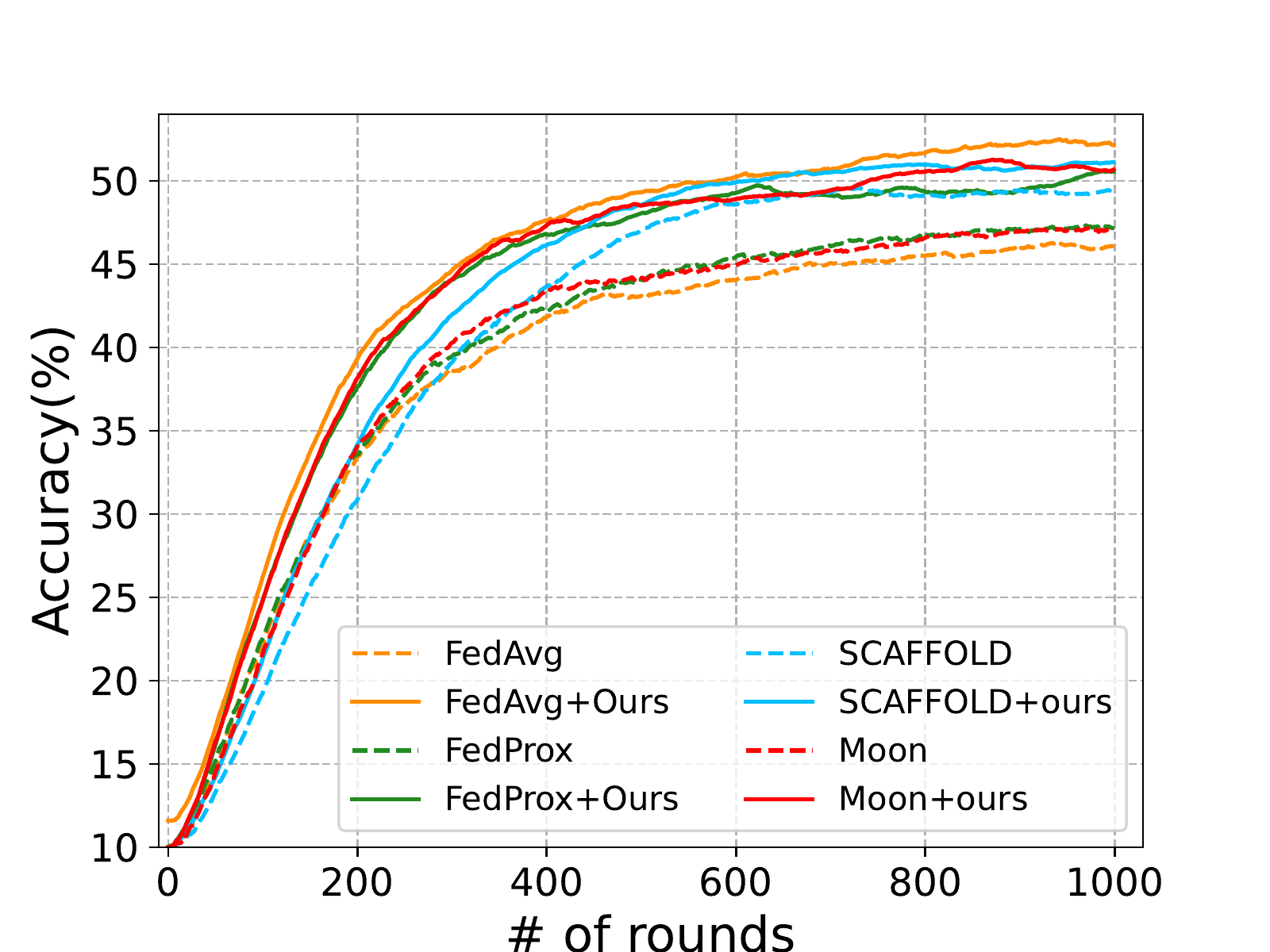}}\\
\vspace{-0.15in}
\subfloat[CIFAR-100 with case 1]{\includegraphics[width=.33\linewidth]{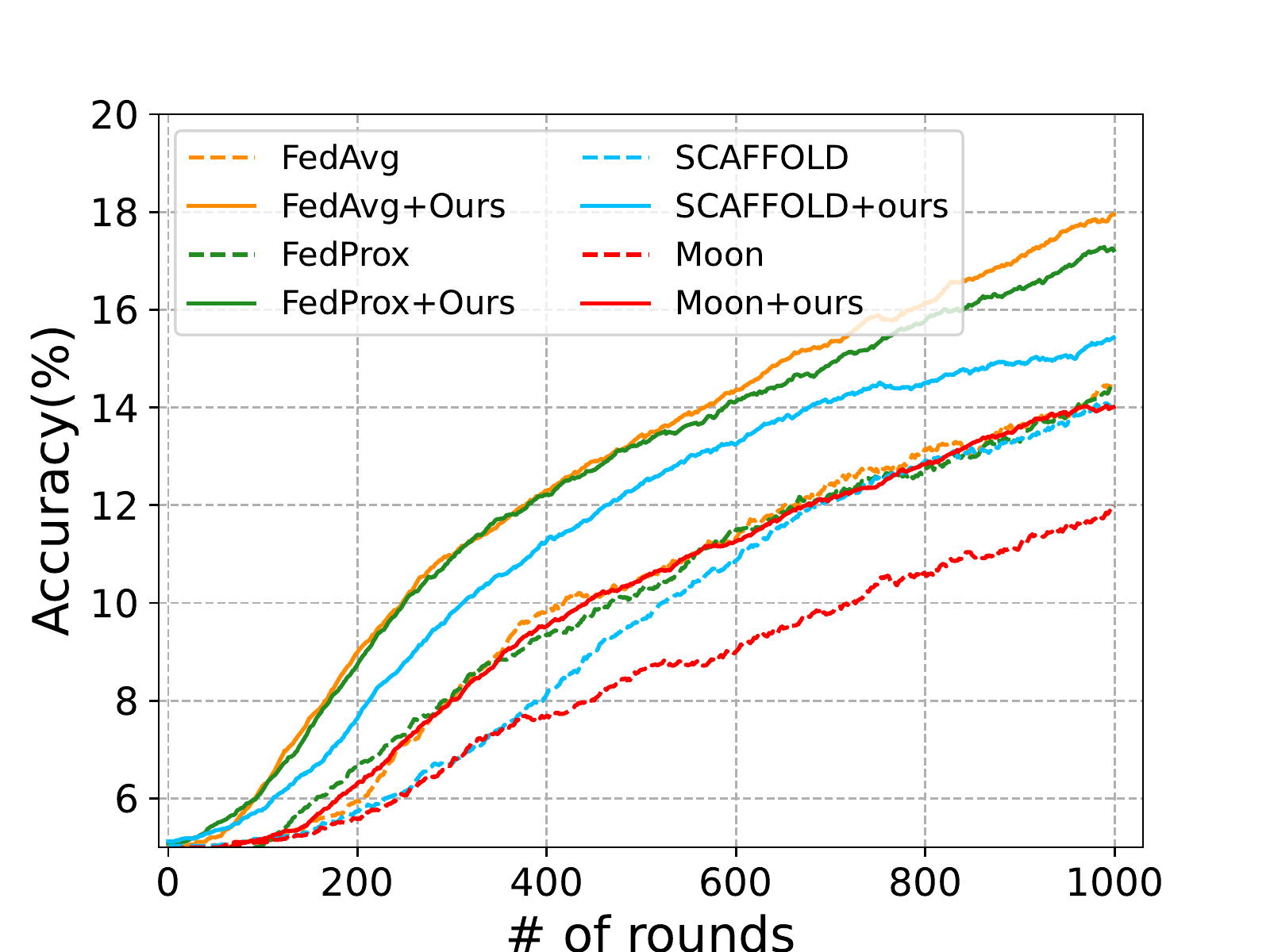}}
\subfloat[CIFAR-100 with case 2]{\includegraphics[width=.33\linewidth]{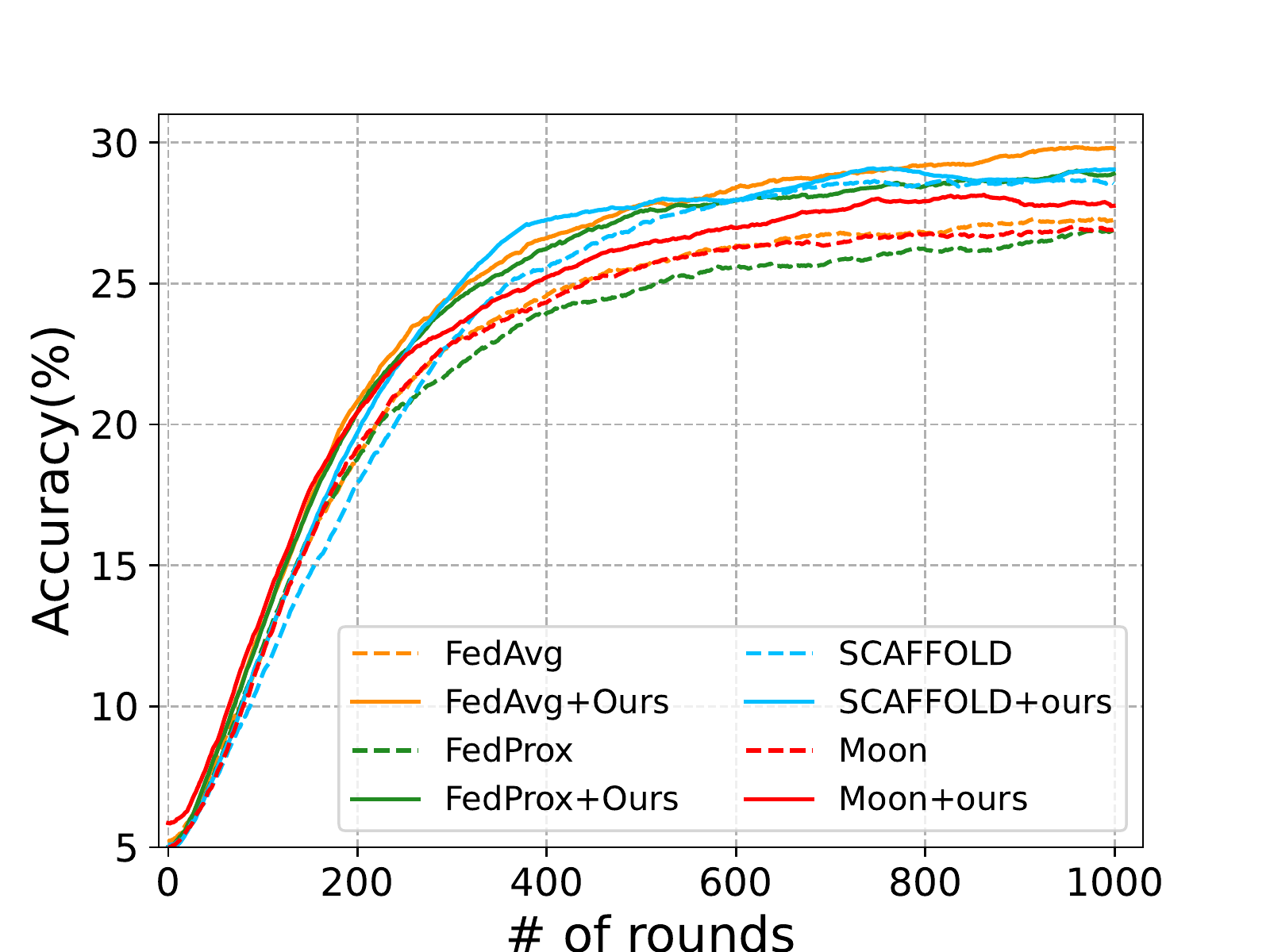}}
\subfloat[CIFAR-100 with case 3]{\includegraphics[width=.33\linewidth]{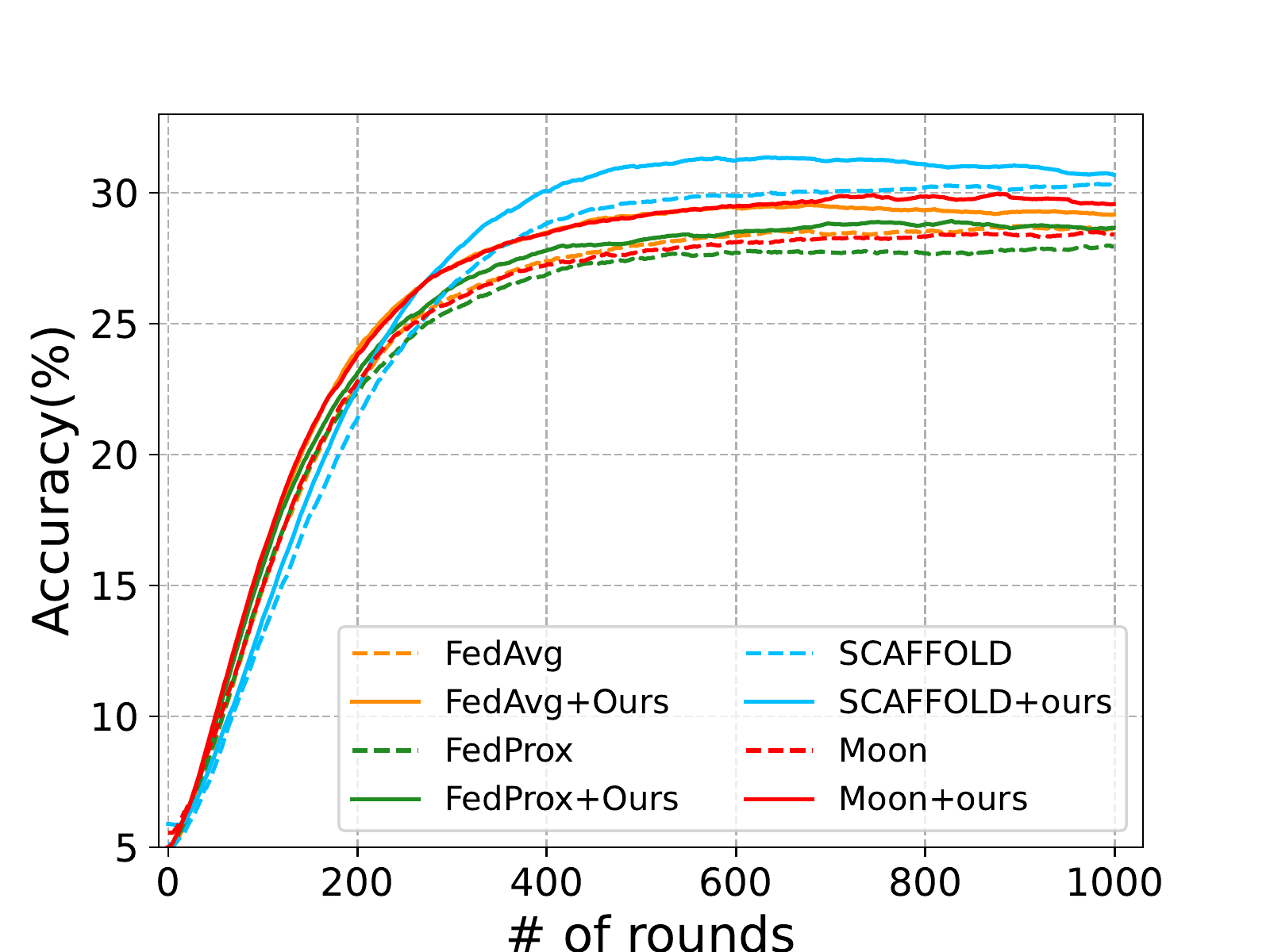}}\\
\vspace{-0.15in}
\subfloat[CINIC-10 with case 1]{\includegraphics[width=.33\linewidth]{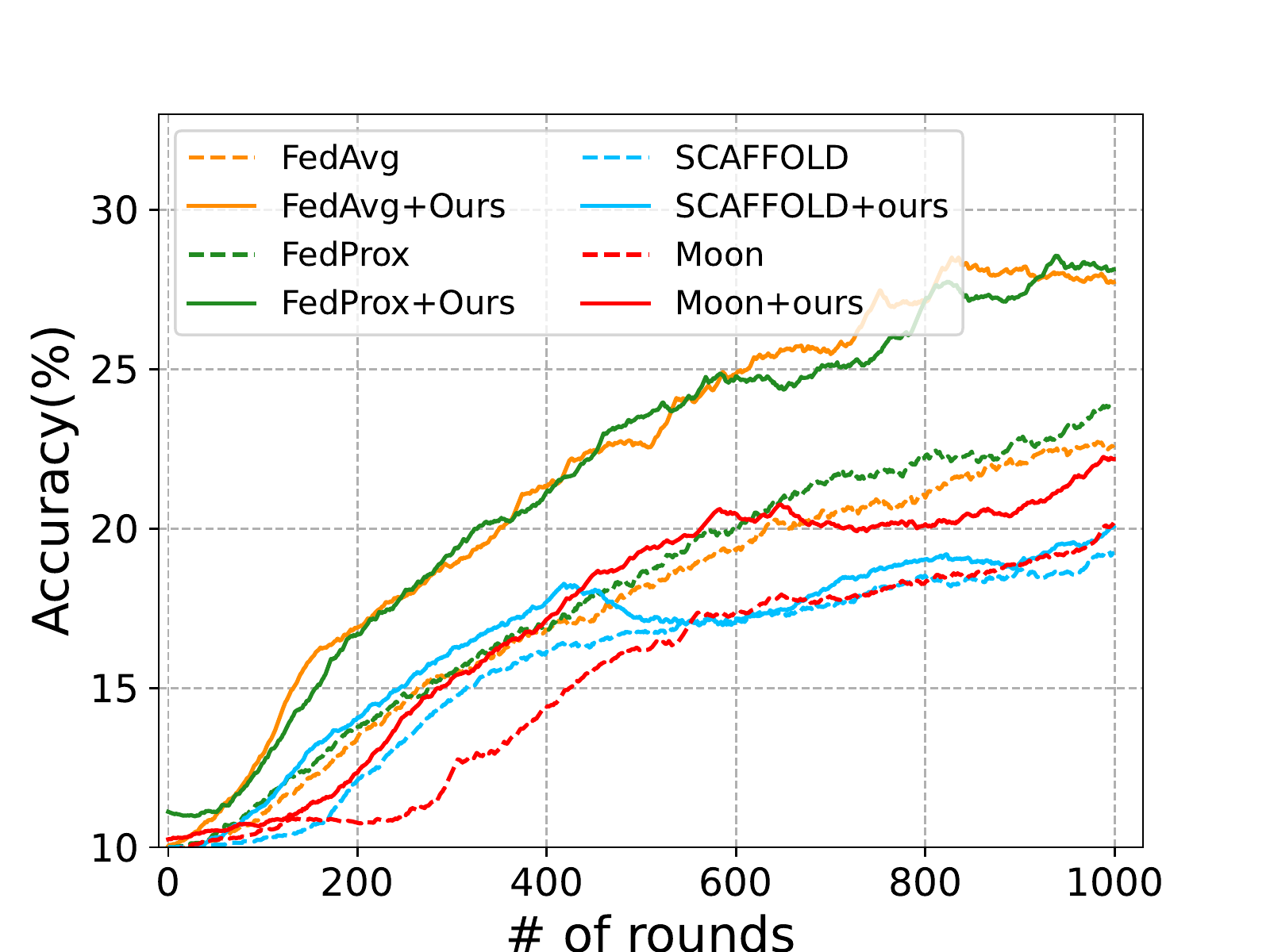}}
\subfloat[CINIC-10 with case 2]{\includegraphics[width=.33\linewidth]{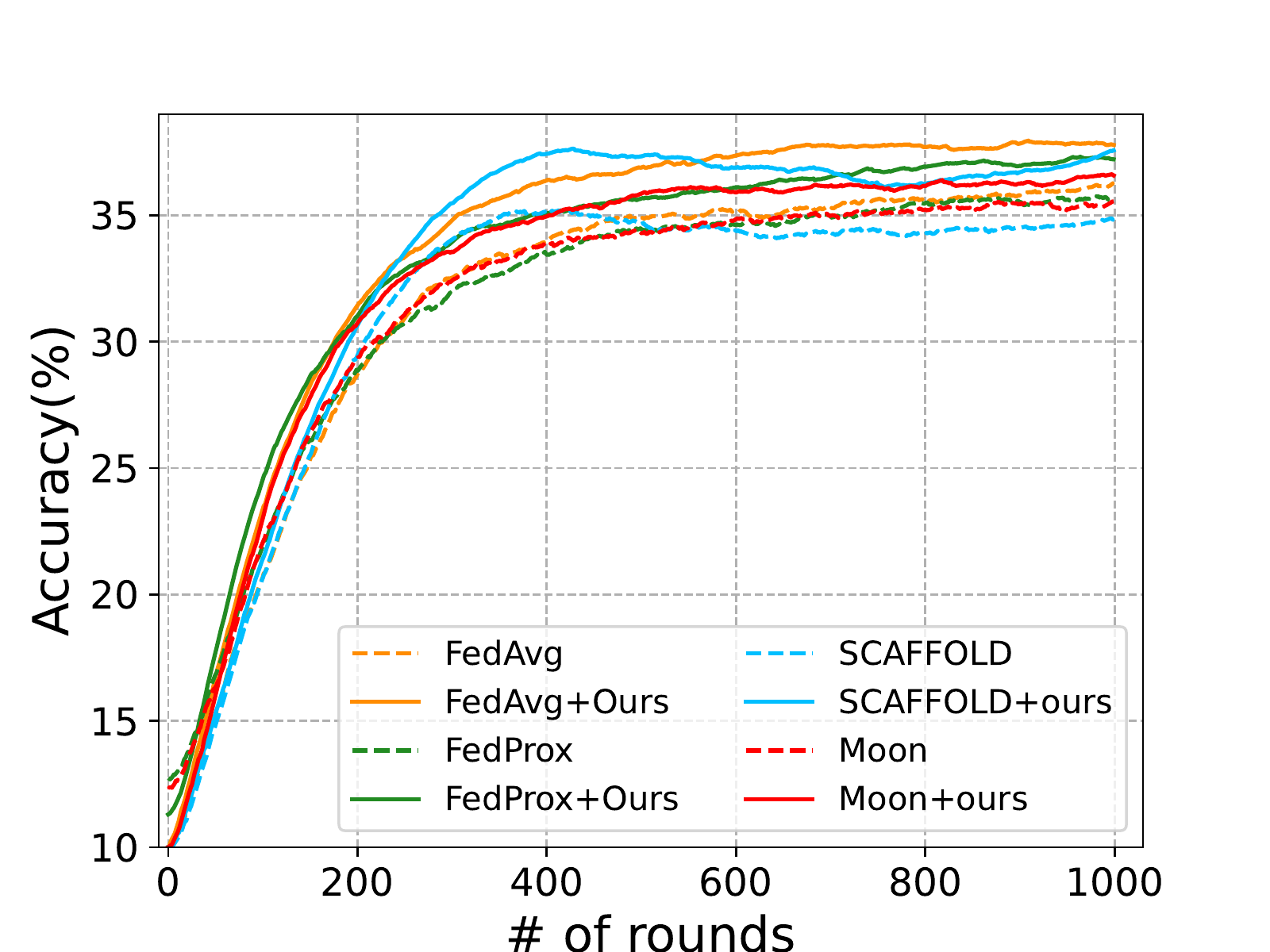}}
\subfloat[CINIC-10 with case 3]{\includegraphics[width=.33\linewidth]{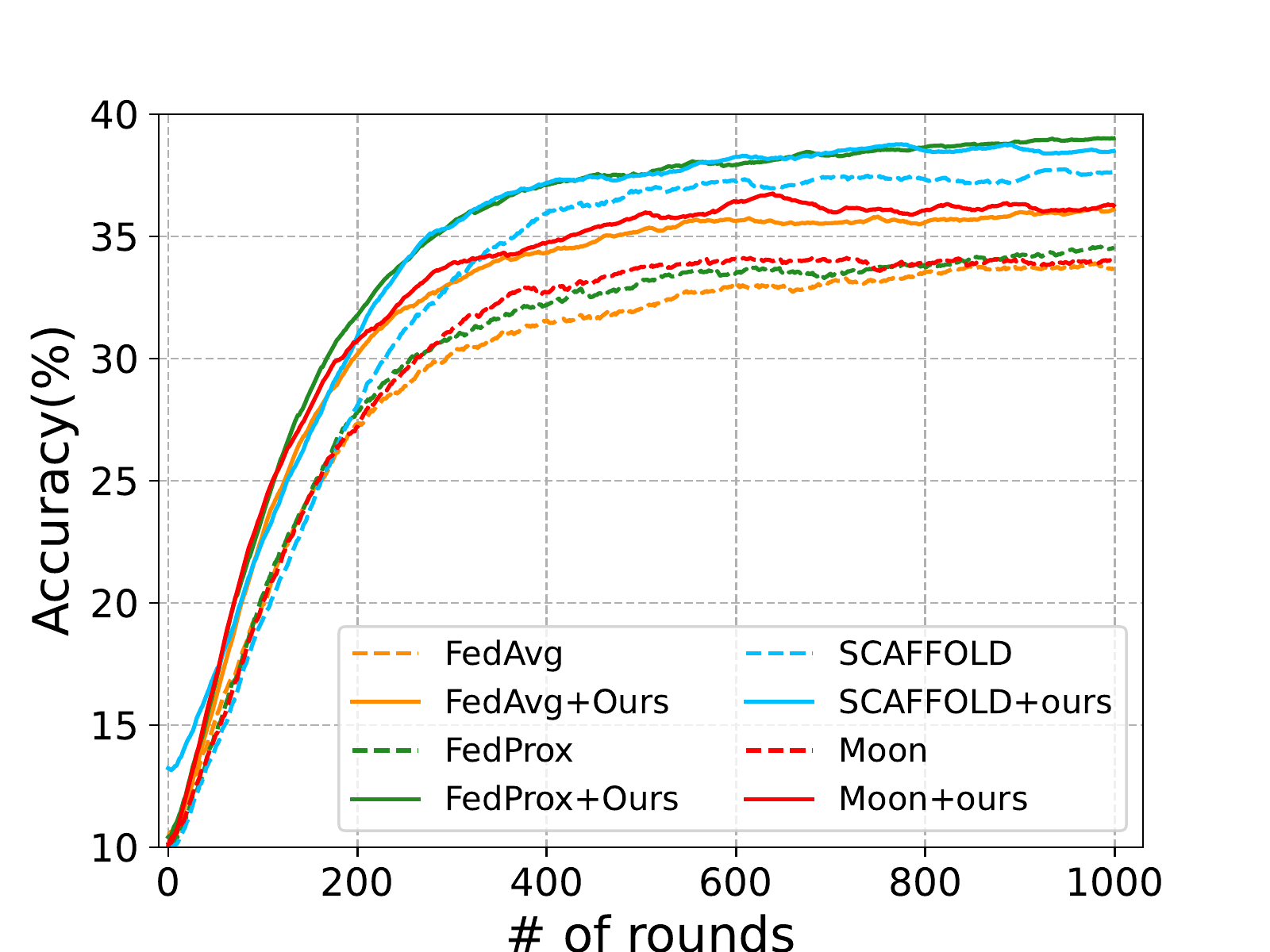}}
\caption{Evaluation of synergy between FedEntropy  and SOTA FL methods.}
\label{f6}
\end{figure*}

\noindent\textbf{Synergy between  FedEntropy  and SOTA FL Methods.}
To evaluate whether our approach can be applied on state-of-the-art methods, 
Figure \ref{f6} presents the accuracy information, where the notation ``{\it X+Ours}'' indicates that a
combination of FL method {\it X} and our approach. We can find that all the four state-of-the-art FL methods 
get notable benefits from our approach.

\begin{table*}[th]
  \caption{Number of  communication rounds to reach target test accuracy on CIFAR-100}
  \label{t6}
  \centering
  \begin{tabular}{c | c c c}
    \hline
    &\multicolumn {3}{c}{Communication Rounds}\\ \cline{2-4}
    &{case 1(acc = 12\%)} &{case 2(acc = 28\%)} &{case 3(acc = 30\%)} \\
    \hline\hline
    {FedAvg} &206.67$\pm$11.02 &233.67$\pm$15.50 &155.33$\pm$17.62 \\
    {FedProx} &222.67$\pm$18.50 &270.67$\pm$61.58 &160.00$\pm$6.58  \\
    {SCAFFOLD} &305.00$\pm$88.71 &246.33$\pm$25.42  &291.33$\pm$15.89  \\
    {Moon} &510.67$\pm$25.01 &219.67$\pm$69.12 &183.00$\pm$17.35 \\
    {Ours} &\textbf{159.67}$\pm$21.46 &\textbf{173.33}$\pm$23.07 &\textbf{117.33}$\pm$4.04 \\
    \hline
  \end{tabular}
\end{table*}

\end{document}